\documentclass[11pt]{article}
\usepackage{acl}
\usepackage{times}
\usepackage{latexsym}
\usepackage[T1]{fontenc}
\usepackage[utf8]{inputenc}
\usepackage{microtype}
\usepackage{inconsolata}
\usepackage{amsmath,amssymb,amsthm,mathtools}
\usepackage{graphicx}
\usepackage{booktabs,multirow,subcaption,float}
\usepackage{xcolor,colortbl}
\usepackage{algorithm,algorithmic}
\usepackage{hyperref,cleveref}
\usepackage{enumitem}
\usepackage{tcolorbox}
\usepackage{adjustbox} 
\usepackage{tabularx}
\usepackage{graphicx}
\graphicspath{{figures/}{latex/figures/}}   % 可以列多个候选目录

\newtheorem{definition}{Definition}
\newtheorem{theorem}{Theorem}

\newtheorem{remark}{Remark}

\newcommand{\method}{\textsc{ERR+}}
\newcommand{\methodbase}{\textsc{ERR}}
\newcommand{\methodlen}{\textsc{RRER}}
\newcommand{\entropy}{\mathcal{H}}
\newcommand{\dH}{\Delta\mathcal{H}}
\newcommand{\ErrScore}{\mathrm{ERR}}

\usepackage{amsmath}

\newcolumntype{Y}{>{\centering\arraybackslash}X}

\title{ERR+: Sequential Entropy Resolution for Efficient and Decisive LLM Reasoning}

\author{%
\large\bfseries
Xin~Jiang$^{1}$\thanks{Equal contribution.}
\quad Minhao~Wang$^{1*}$
\quad Wen~Wu$^{1}$\thanks{Corresponding author.} \\
\large\bfseries
\quad Zhentao~Xie$^{1}$ 
\quad Shangheng~Du$^{1}$ 
\quad Jinxin~Shi$^{1}$ 
\quad Jiabao~Zhao$^{1}$ 
\quad Liang~He$^{1}$ 
\quad Weicong~Chen$^{2}$ \\
$^{1}$East China Normal University \\
$^{2}$ByteDance \\
\tt \{51275901099, 51275901104\}@stu.ecnu.edu.cn, wwu@cs.ecnu.edu.cn
}

\begin{document}
\maketitle

% ============================================================
\begin{abstract}
Large reasoning models achieve strong performance on complex tasks by
generating extended chain-of-thought (CoT) traces via reinforcement
learning with verifiable rewards (RLVR). While current RLVR methods
have achieved strong results with correctness-based reward signals,
they provide limited guidance on the quality of the reasoning process
itself, leaving the internal reasoning structure largely unoptimized.
Through empirical analysis across multiple model families, we identify
a consistent pattern: correct reasoning trac    es exhibit more frequent
and larger token-level entropy drops within the thinking phase than
incorrect ones. We propose \method{}, a
two-phase RLVR framework grounded in this observation. The first phase
trains with the Entropy Relief Reward (\methodbase{}), a bonus
proportional to cumulative token-level entropy drops in the thinking
phase, log-normalized by response length. Unlike prior methods that
suppress entropy, \methodbase{} rewards the resolution of uncertainty
while leaving exploratory high-entropy states unconstrained. The
second phase introduces the Robust Relative Efficiency Reward, which scores each response's length against
co-generated peers via a $\tanh$-transformed within-group $z$-score.
We provide a formal analysis showing that joint optimization of the
two objectives induces gradient conflict in early training, motivating
the sequential design (Theorem~\ref{thm:conflict}). Experiments on
five datasets demonstrate consistent improvements in both accuracy and response conciseness across model
backbones. Our code is available at \url{https://github.com/XrkArul/err_response}.
\end{abstract}
% ============================================================

\begin{figure*}[t]
  \centering
  \begin{subfigure}[t]{0.48\textwidth}
    \centering
    \includegraphics[width=\linewidth]{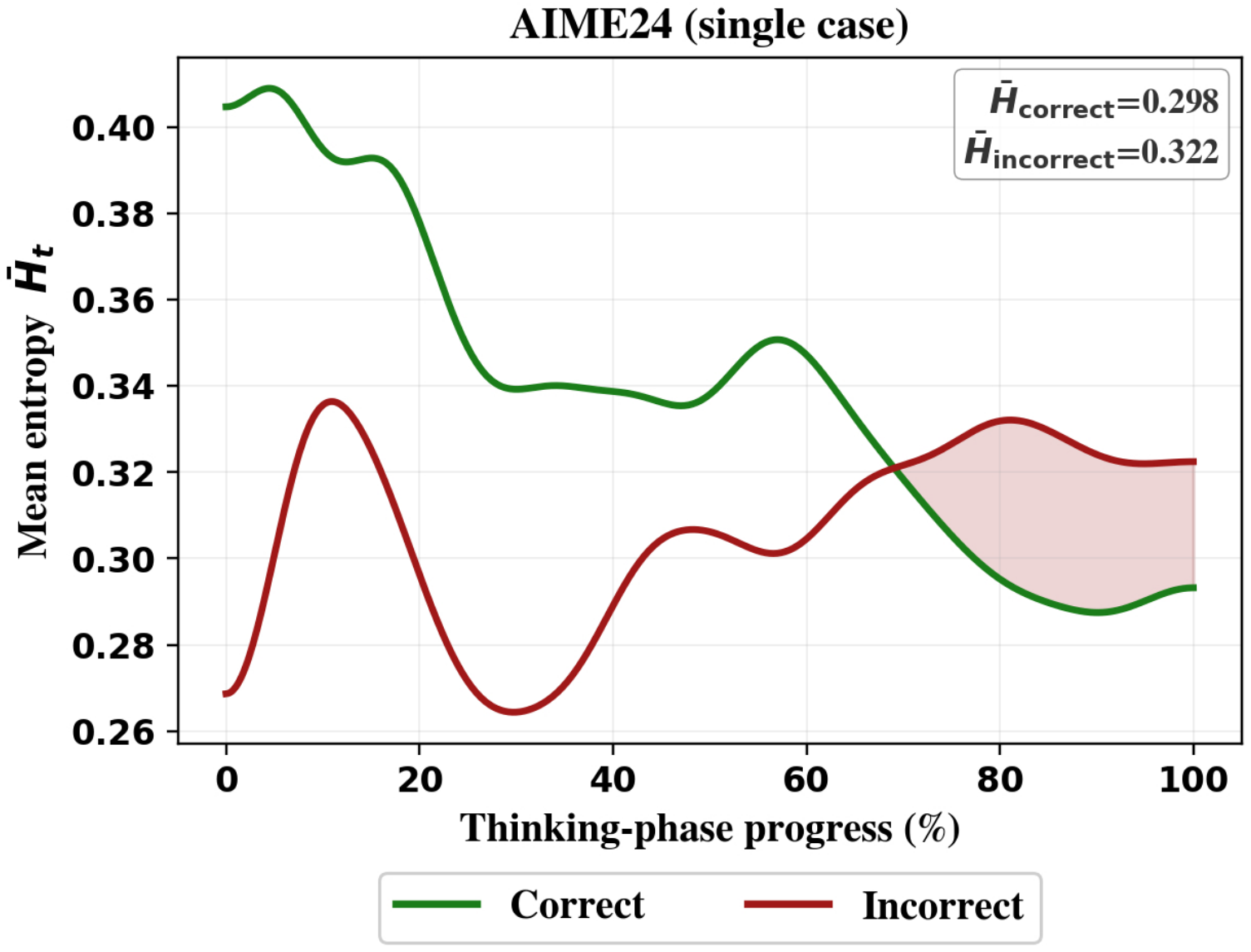}
    \caption{Single AIME24 case. Running-mean entropy of one correct
    (blue) and one incorrect (red) rollout from the same problem and
    model. The correct trace commits decisively after an exploratory
    phase; the incorrect trace remains at persistently high entropy.
    Curves are Gaussian-smoothed
    ($\sigma = \max(40,\lfloor T/25\rfloor)$).}
    \label{fig:entropy_case}
  \end{subfigure}
  \hfill
  \begin{subfigure}[t]{0.48\textwidth}
    \centering
    \includegraphics[width=\linewidth]{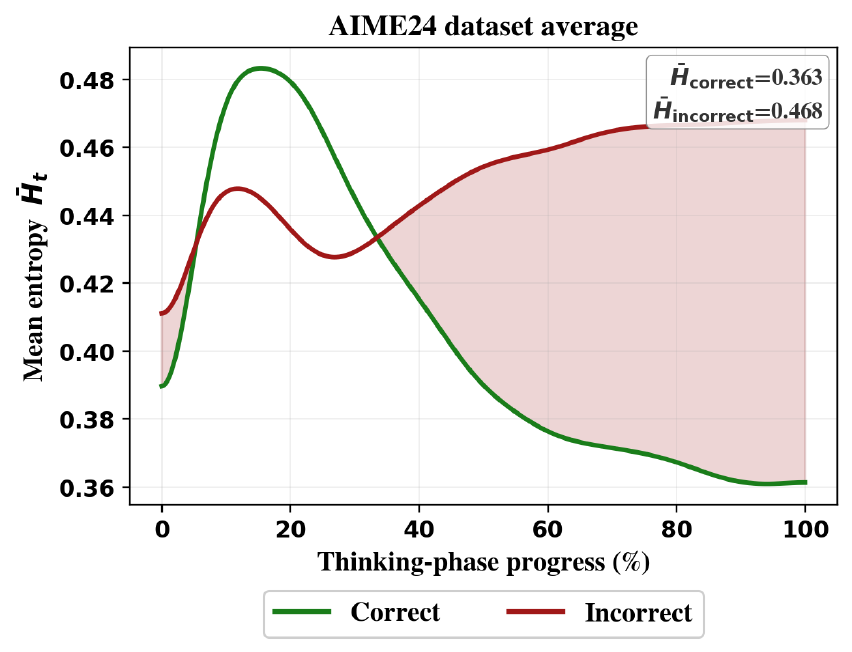}
    \caption{Dataset-level aggregation via bootstrap sampling. We draw 
    16 independent samples from all AIME24 rollouts, separate them by 
    correctness, compute smoothed running-mean entropy for each group, 
    and average across samples. The resulting traces confirm that correct 
    rollouts exhibit more decisive entropy drops than incorrect ones 
    ($\sigma = 40$, resampled to 501 points).}
    \label{fig:entropy_avg}
  \end{subfigure}
  \caption{Running-mean entropy trajectories for correct (blue) vs.\ 
  incorrect (red) rollouts on AIME24. Both at the single-case and 
  dataset-aggregated level, correct traces exhibit more decisive 
  entropy drops than incorrect ones.}
  \label{fig:entropy_both}
\end{figure*}

\section{Introduction}
\label{sec:intro}

Large language models (LLMs) have demonstrated remarkable reasoning
capabilities when employing chain-of-thought (CoT)
prompting~\citep{wei2022chain}. Recent large reasoning
models~\citep{deepseek-r1,openai-o1,qwen3} build on this by
explicitly encouraging an extended thinking phase via special
\texttt{<think>}\ldots\texttt{</think>} tokens before the final
answer, and are trained with RLVR algorithms such as
GRPO~\citep{shao2024deepseekmath} and DAPO~\citep{yu2025dapo}.

A structural limitation of this paradigm becomes pronounced in
high-accuracy regimes. Within any sampled group, multiple responses
may reach the correct answer through qualitatively different
reasoning paths. Under within-group normalization, the advantage of
response $i$ is
$A^{(i)} = (R(y^{(i)}) - \bar{R}) / (\sigma_R + \delta)$; when all
correct responses receive identical binary rewards, the
within-correct-group variance collapses to zero. The policy thus
receives limited signal to prefer concise, well-structured reasoning
over verbose reasoning of equal correctness.

Recent work has established token-level entropy as an informative
internal signal for reasoning
quality~\citep{wang2025beyond,chen2025unreasonable,
tan2025gtpo,huang2025pear}. High-entropy tokens correspond to
exploratory branch points; low-entropy tokens reflect deterministic
generation. Critically, \citet{wang2025beyond} show that the top-20\%
highest-entropy tokens account for the majority of RLVR training
gains, identifying them as the primary reasoning junctions. Existing
entropy-based methods, however, operate on absolute entropy levels:
minimizing entropy globally~\citep{chen2025unreasonable}, scaling
gradients by entropy magnitude~\citep{tan2025gtpo}, or penalizing
thinking-phase entropy as a length proxy~\citep{huang2025pear}.
Penalizing high entropy directly suppresses the exploratory positions
that \citet{wang2025beyond} identify as the locus of RLVR learning;
accordingly, PEAR~\citep{huang2025pear} reports consistent accuracy
degradation alongside its length reductions.

We take a different perspective: rather than operating on entropy
levels, we focus on entropy \emph{changes}. We observe empirically
that correct reasoning traces exhibit significantly more frequent
and deeper token-level entropy drops within the thinking phase than
incorrect ones. This suggests that
rewarding entropy decreases, not penalizing entropy presence,
provides a quality signal that leaves exploration unconstrained.

We propose \method{}, a two-phase RLVR framework. The first phase
trains with \methodbase{}, which augments the correctness reward with
a bonus proportional to cumulative entropy drops, log-normalized by
response length. The second phase refines the model using \methodlen{},
a difficulty-aware length signal based on within-group $z$-score
normalization with $\tanh$ saturation. The separation of phases is not
merely heuristic: we provide a formal gradient-conflict analysis
(Theorem~\ref{thm:conflict}, proof in Appendix~\ref{app:proof})
showing that the two reward components produce conflicting policy
gradient directions in early training, and that this conflict resolves
at first-phase convergence, making sequential training preferable to
joint optimization.

Our contributions are as follows:
\begin{itemize}[leftmargin=*, noitemsep=3pt, topsep=6pt]
  \item We establish empirically that correct reasoning traces exhibit
    more frequent and deeper token-level entropy drops than incorrect
    ones, consistently across model families and reasoning tasks.

  \item We propose \method{}, translating this finding into two
    complementary reward signals: \methodbase{} rewards entropy
    resolution without suppressing high-entropy exploration, and
    \methodlen{} provides a prompt-difficulty-aware length signal
    robust to group outliers.

  \item We formally characterize the gradient conflict between the two
    objectives and prove that sequential training is preferable to
    joint optimization in early training regimes.

  \item Evaluations on five benchmarks show consistent accuracy
    improvements and response length reductions over baselines across
    four model backbones.
\end{itemize}

\begin{figure*}[t]
  \centering
  \includegraphics[width=\textwidth]{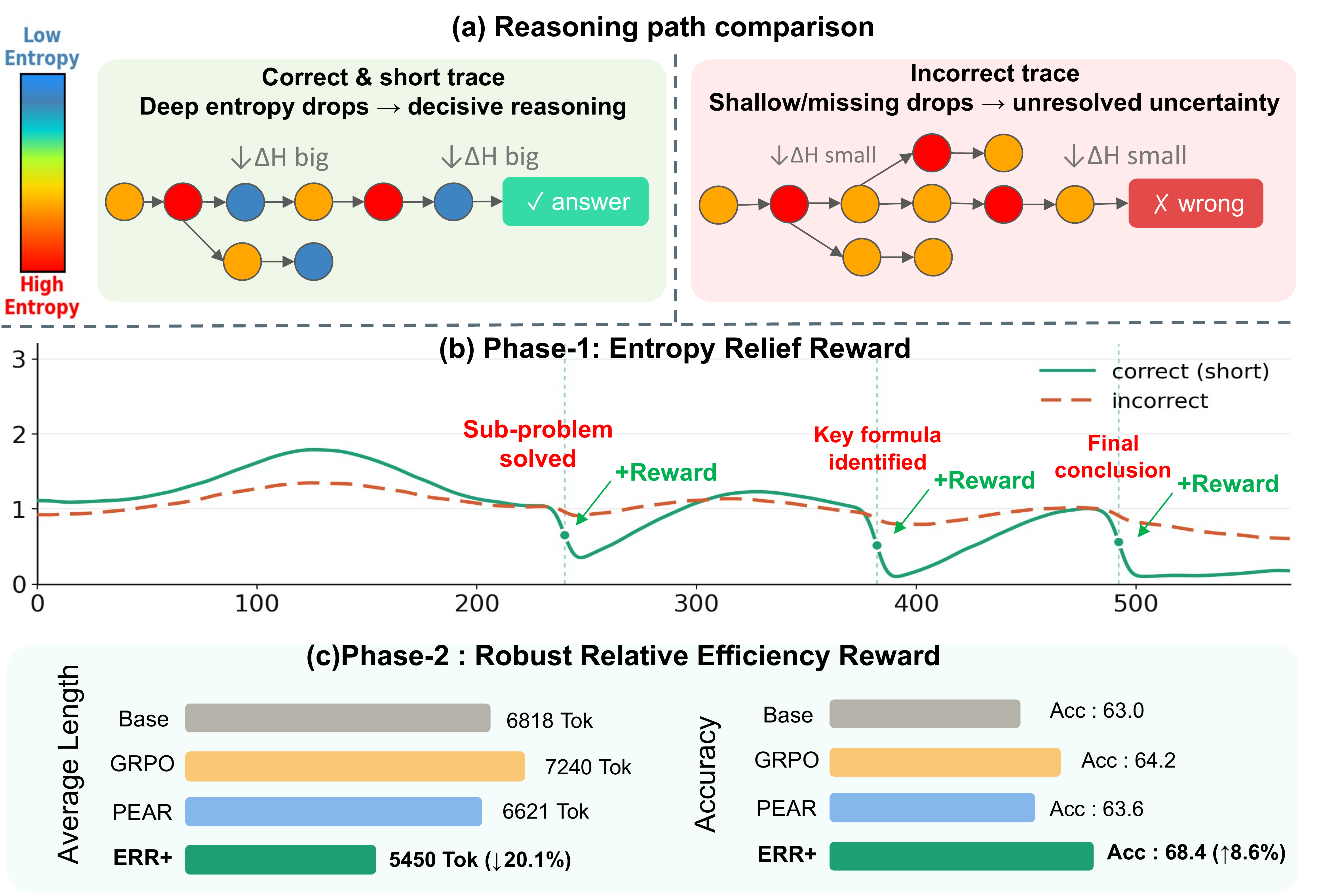}
  \vspace{0.5em}
  \caption{%
    \textbf{Overview of \method{}.}
    \textbf{(a)} In a correct short trace, high-entropy tokens are
    consistently followed by large entropy drops, signaling decisive
    commitment; in an incorrect trace, drops are shallow or absent.
    \textbf{(b) Phase 1 (\methodbase{}).}
    The correct trace (solid green) shows three pronounced drops at
    reasoning milestones (sub-problem solved, key formula identified,
    final conclusion), each yielding a \textbf{+Reward} bonus;
    the incorrect trace (dashed orange) remains flat.
    \textbf{(c) Phase 2 (\methodlen{}).}
    On DS-R1-1.5B, \method{} reduces response length by 20\%
    and improves accuracy by 8.7\%, decoupling the
    accuracy--conciseness trade-off present in all baselines.%
  }
  \label{fig:overview}
\end{figure*}

% ============================================================
\section{Related Work}
\label{sec:related}

\subsection{Reinforcement Learning for LLM Reasoning}

RLVR has become the dominant training paradigm for large reasoning
models~\citep{deepseek-r1,openai-o1,qwen3,kimi15}.
GRPO~\citep{shao2024deepseekmath} removes the critic via within-group
advantage normalization. DAPO~\citep{yu2025dapo} addresses gradient
dilution and entropy collapse with dynamic sampling, token-level
policy gradient loss, and clip-higher. VAPO~\citep{yu2025vapo}
decouples value estimation across token types;
Dr.~GRPO~\citep{liu2025drgrpo} corrects a length bias in GRPO
normalization; GDPO~\citep{liu2025gdpo} extends to multi-objective
settings. For length efficiency, Kimi~k1.5~\citep{kimi15} normalizes
penalties by group extremes, LASER~\citep{luo2025laser} uses
fixed-budget step-function penalties, and ConMax~\citep{hu2025conmax}
proxies conciseness with answer-phase confidence. All these length
signals apply equal pressure regardless of problem difficulty.
\method{} addresses this via within-group $z$-score normalization,
using the co-generated group mean as a difficulty proxy.

\subsection{Reasoning Through Entropy Control}

Token-level entropy has emerged as a key lens for understanding and
improving RLVR. \citet{cui2025entropy} show that entropy collapse at
high-entropy positions correlates with performance saturation;
\citet{zhang2025nofree} establish that these tokens receive
disproportionately large gradient updates and suppressing them
degrades accuracy. \citet{wang2025beyond} formalize the asymmetry:
training on only the top-20\% highest-entropy ``forking tokens''
matches or exceeds full-token RLVR, while the low-entropy majority
is largely ineffective. Building on these insights,
\citet{chen2025unreasonable} show entropy minimization alone improves
reasoning accuracy; \citet{tan2025gtpo} weight gradients by token
entropy; \citet{zhang2025edge} add entropy-based exploration bonuses;
\citet{cheng2025reasoning} use an entropy coefficient for diverse
solution paths. PEAR~\citep{huang2025pear} penalizes average
thinking-phase entropy as a length proxy, but degrades accuracy by
suppressing the exploratory positions \citet{wang2025beyond} identify
as critical. TokenSqueeze~\citep{tokensqueeze2025} compresses
reasoning traces by selectively dropping low-importance tokens, but
similarly struggles to maintain accuracy while reducing length.
A common limitation across these methods is the difficulty of
simultaneously improving accuracy and reducing response length:
optimizing for one objective tends to come at the cost of the other.
\method{} addresses this by operating on entropy \emph{changes}:
\methodbase{} rewards uncertainty resolution at each step without
constraining exploration, enabling \method{} to improve both
accuracy and conciseness jointly.

% ============================================================
\section{Preliminary Analysis}
\label{sec:prelim}

We study token-level entropy dynamics in reasoning traces across
multiple large reasoning models on various mathematical reasoning
benchmarks. Let $x$ be the input query and $y=(y_1,\ldots,y_T)$ a response from
$\pi_\theta$, with $T_k$ the index of \texttt{</think>}
($T_k=T$ if absent). The token-level entropy at thinking-phase
position $t$ is:
\begin{equation}
  \small
  \entropy_t = -\sum_{v \in \mathcal{V}}
  \pi_\theta\!\left(v \mid x, y_{<t}\right)
  \log \pi_\theta\!\left(v \mid x, y_{<t}\right),
  \label{eq:prelim_entropy}
\end{equation}
computed from per-token logits at no additional inference cost.
We define the per-step entropy change
$\dH_t = \entropy_t - \entropy_{t-1}$ ($t\geq 2$), with $\dH_t<0$
indicating a local drop. Responses are partitioned into correct
($\mathcal{C}$) and incorrect ($\mathcal{W}$).

\paragraph{Entropy drops correlate with correctness.}
Figures~\ref{fig:entropy_case} and~\ref{fig:entropy_avg} examine
this pattern through running-mean entropy traces, providing a 
direct view of how entropy evolves over the course of reasoning.

Figure~\ref{fig:entropy_case} shows a single AIME24 problem
with one correct and one incorrect rollout drawn from the same
model. The correct trace exhibits a pronounced and sustained
decline in running-mean entropy after an initial exploratory
phase, signaling decisive commitment to a solution path; the
incorrect trace remains persistently high throughout, reflecting
unresolved uncertainty.

Figure~\ref{fig:entropy_avg} generalizes this comparison to the
dataset level. Rather than selecting representative instances 
per problem, we draw 16 independent bootstrap samples from the 
full set of AIME24 rollouts. For each sample, we separate the 
trajectories into correct and incorrect groups based on final 
answer accuracy, compute the smoothed running-mean entropy curve 
for each group, and then average these curves across the 16 
samples. This aggregation strategy ensures that the resulting 
traces reflect population-level trends rather than artifacts of 
problem-specific selection. The pooled correct traces consistently 
exhibit sharper entropy declines compared to their incorrect 
counterparts, confirming that the entropy-drop signature is a 
robust statistical signal across the dataset.

Together, these observations establish entropy drops as a
reliable indicator of reasoning quality, directly motivating
the reward design in Section~\ref{sec:method}.

% ============================================================
\section{Method}
\label{sec:method}

Section~\ref{sec:prelim} establishes that entropy drops in the
thinking phase are a reliable signal for reasoning quality: correct
traces exhibit more frequent and larger drops than incorrect ones.
We translate this observation into two reward components.
\methodbase{} rewards cumulative entropy drops to improve reasoning
quality. \methodlen{} evaluates response length relative to
co-generated peers to encourage conciseness. We formally show in
Theorem~\ref{thm:conflict} that optimizing these two objectives
jointly in early training produces gradient conflict, motivating
a sequential training schedule. The full pipeline is illustrated
in Figure~\ref{fig:overview}.

We denote by $y^*$ the ground-truth answer, by
$\mathcal{G} = \{y^{(1)}, \ldots, y^{(G)}\}$ the group of $G$
responses sampled for query $x$, and use $\entropy_t$ as defined
in Eq.~\eqref{eq:prelim_entropy}.

\subsection{Entropy Relief Reward}
\label{sec:err}

Section~\ref{sec:prelim} shows that entropy drops within the
thinking phase reliably distinguish correct traces from incorrect
ones. A natural reward is to accumulate the magnitudes of these
drops. However, a raw sum $\sum_t r_t$ would grow with thinking
length, creating an incentive to generate longer traces purely to
accumulate more drops regardless of their quality. We address this
by normalizing by $\log(T_k+1)$, which grows sub-linearly in $T_k$
and therefore imposes a diminishing per-token density requirement
as traces lengthen.

\begin{definition}[Entropy Relief Score]
\label{def:err}
The per-token entropy relief at position $t \in [2, T_k]$ is:
\begin{equation}
  \small
  r_t = \max\!\left(\entropy_{t-1} - \entropy_t - \epsilon,\; 0\right),
  \label{eq:relief}
\end{equation}
where $\epsilon \geq 0$ suppresses noise-level fluctuations. The
\methodbase{} score is:
\begin{equation}
  \small
  \ErrScore(y) =
  \frac{\displaystyle\sum_{t=2}^{T_k} r_t}{\log\!\left(T_k + 1\right)}.
  \label{eq:err_score}
\end{equation}
\end{definition}

We prefer $\log(T_k+1)$ over linear normalization $(\sum r_t)/T_k$
because a fixed per-token drop density target would unfairly
penalize hard problems requiring genuinely extended reasoning. The
first-phase reward is:
\begin{equation}
  \small
  R_1(y, y^*) =
  \begin{cases}
    \min\!\left(R_b + \lambda\,\ErrScore(y),\; R_{\max}\right)
      & \text{correct,}\\[3pt]
    R_f & \text{incorrect,}\\[3pt]
    0   & \text{unparseable,}
  \end{cases}
  \label{eq:reward1}
\end{equation}
where $R_b$ is the base correctness reward, $R_f$ a small format
reward, and $R_{\max}$ caps the total reward to prevent any single
response from dominating group-level advantage. The \methodbase{} bonus is gated on correctness to prevent
the model from exploiting entropy-drop scores by producing
structurally plausible but factually wrong traces.
Importantly, \methodbase{} does not claim that every entropy
drop corresponds to a semantically correct intermediate step,
nor does it replace step-level verification. Its role is
to supply within-correct-group process signal when only
final-answer verification is available: a trace receives
the bonus only if the final answer is correct, so entropy
drops in ultimately wrong traces are never reinforced.
Entropy drops should therefore be interpreted as
\emph{correctness-gated commitment events} rather than
standalone semantic validators.

\subsection{Robust Relative Efficiency Reward}
\label{sec:rrer}

We design \methodlen{} to penalize unnecessarily long responses
without applying uniform length pressure regardless of problem
difficulty. Harder prompts naturally elicit longer responses, so an
absolute length penalty would over-penalize hard problems. We
address this by evaluating each response relative to its
co-generated peers: the within-group mean $\mu_L$ acts as an
implicit difficulty proxy without requiring any external label.

We standardize response length within the group:
\begin{equation}
  \small
  z_i = \frac{L_i - \mu_L}{\sigma_L + \varepsilon},
  \label{eq:zscore}
\end{equation}
where $\mu_L$, $\sigma_L$ are the group mean and standard deviation,
and $\varepsilon = 10^{-5}$ prevents division by zero. The efficiency
score maps $z_i$ through:
\begin{equation}
  \small
  \lambda_i = \tanh\!\left(-\gamma\, z_i\right),
  \label{eq:lambda}
\end{equation}
with $\gamma > 0$ controlling sensitivity. We use $\tanh$ rather
than a linear mapping to ensure robustness to outliers: if one
response hits the maximum token limit, a linear scheme collapses
reward differentials for all other responses, whereas $\tanh$
bounds $\lambda_i \in (-1, 1)$ regardless of the group's length
distribution.

To prevent short but incorrect responses from receiving length
bonuses, we apply asymmetric allocation:
\begin{equation}
  \small
  f(\lambda_i, c_i) =
  \begin{cases}
    \lambda_i                        & c_i = 1,\\[2pt]
    \min\!\left(0,\,\lambda_i\right) & c_i = 0,
  \end{cases}
  \label{eq:asymmetric}
\end{equation}
where $c_i \in \{0,1\}$ is correctness. When $\sigma_L < \varepsilon$
all group responses have equal length and $\lambda_i = 0$,
deactivating \methodlen{}. The second-phase reward is:
\begin{equation}
  \small
  R_2(y, y^*) =
  \begin{cases}
    R_b + \alpha\,\lambda_i
      & \text{correct,}\\[4pt]
    R_f + \alpha\,\min\!\left(0,\lambda_i\right)
      & \text{otherwise.}
  \end{cases}
  \label{eq:reward2}
\end{equation}
Since $|\alpha\lambda_i| < \alpha$, the reward range is naturally
bounded without an explicit cap.

\subsection{Sequential Training via Gradient-Conflict Analysis}
\label{sec:theory}

We now show formally why \methodbase{} and \methodlen{} must be
applied sequentially. Let $\mathcal{L}_k(\theta) =
-\mathbb{E}_{\pi_\theta}[R_k]$ be the expected negative reward for
phase $k \in \{1,2\}$, and let
$g_k(\theta) = \nabla_\theta \mathcal{L}_k(\theta)$.
Let $\bar{\entropy} = \mathbb{E}_{y \sim \pi_\theta,\, t \in
\mathcal{T}(y)}[\entropy_t]$ be the mean thinking-phase entropy.
We define the \emph{mid-exploration set}
$\mathcal{T}_\rho(y) = \{t \in \mathcal{T}(y) :
\entropy_t > \bar{\entropy},\, r_t = 0\}$
as the set of high-entropy positions not yet followed by a committed
drop, the mid-exploration fraction
$\rho(\theta) = \mathbb{E}_{y \sim \pi_\theta}[|\mathcal{T}_\rho(y)|
/ |\mathcal{T}(y)|]$,
and the entropy surplus
$\Delta\bar{\entropy} = \mathbb{E}[\entropy_t - \bar{\entropy}
\mid t \in \mathcal{T}_\rho] > 0$.
Let $C = \lambda\,\Delta\bar{\entropy} / \log(T_k+1) > 0$.

\begin{theorem}
\label{thm:conflict}
For any $\theta$ with $\rho(\theta) > 0$:
\begin{equation}
  \small
  \left\langle g_1(\theta),\, g_2(\theta) \right\rangle
  \;\leq\; -C \cdot \rho(\theta),
  \label{eq:conflict}
\end{equation}
and for any joint update $\theta' = \theta - \eta(g_1 + g_2)$:
\begin{equation}
  \small
  \mathcal{L}_1(\theta') \;\geq\;
  \mathcal{L}_1\!\left(\theta - \eta g_1\right)
  + \eta\, C\,\rho(\theta) + O(\eta^2).
  \label{eq:degradation}
\end{equation}
At first-phase convergence $\theta_1^*$, $\rho(\theta_1^*) \approx 0$
and the degradation term $\eta\,C\,\rho(\theta_1^*) \approx 0$.
\end{theorem}

The proof is in Appendix~\ref{app:proof}. Theorem~\ref{thm:conflict} intentionally isolates this one
specific conflict—early length pressure removing high-entropy
exploratory context before it becomes rewarded entropy
relief—rather than fully characterising all GRPO dynamics.
It yields two directly testable predictions:
(i)~joint optimization should be harmful early in training,
which is confirmed by joint $R_1+R_2$ achieving 62.6\%
average accuracy versus \method{}'s 68.4\%; and
(ii)~the conflict should diminish after phase~1 converges,
which is confirmed empirically by the gradient cosine between
$g_1$ and $g_2$ improving from $-0.45$ (before training)
to $-0.05$ (after phase~1) and $+0.02$ (after \method{})—
see Table~\ref{tab:mechanism} in Appendix~\ref{app:structure}. Intuitively, in early
training $g_2$ reduces length by suppressing mid-exploration tokens
(high $\entropy_t$, zero $r_t$), which are precisely the
high-entropy context preceding committed drops; removing them
reduces $r_{t'}$ at the subsequent committed position, directly
harming $\mathcal{L}_1$. After phase-1 converges, the tokens $g_2$
would prune are low-entropy redundant spans with negligible
contribution to $\ErrScore$, so the conflict vanishes.

\subsection{Training Procedure}
\label{sec:procedure}

Algorithm~\ref{alg:training} summarizes \method{}. In phase~1,
$R_1$ is computed independently per response from its entropy
sequence and correctness label. In phase~2, $\mu_L$ and $\sigma_L$
are computed over the full group before individual $\lambda_i$
values are assigned. Both phases apply standard GRPO advantage
normalization:
\begin{equation}
  \small
  A^{(i)} =
  \frac{\displaystyle R\!\left(y^{(i)}, y^*\right)
        - \frac{1}{G}\sum_j R\!\left(y^{(j)}, y^*\right)}
       {\mathrm{std}\!\left(\left\{R\!\left(y^{(j)},
        y^*\right)\right\}\right) + \delta}.
  \label{eq:advantage}
\end{equation}
Under binary rewards, identical correct rewards yield zero
within-group gradient for the correct subset. \method{} restores a
non-degenerate signal: in phase~1 through differing entropy-drop
profiles, and in phase~2 through differing group-relative lengths.
$N_1$ and $N_2$ are the phase-1 and phase-2 checkpoints, each selected by the same rule: among checkpoints whose response length is not an outlier, we take the one with the highest validation accuracy, applied consistently across variants, backbones, and optimizers.

\begin{algorithm}[t]
\caption{\method{} Training}
\label{alg:training}
\begin{algorithmic}[1]
\REQUIRE Base model $\pi_{\theta_0}$; hyperparameters
         $\epsilon, \gamma, \lambda, \alpha, R_b, R_f, R_{\max},
         \varepsilon$; step counts $N_1, N_2$
\STATE \textbf{Phase 1 (\methodbase{}):}
\STATE $\pi_\theta \leftarrow \pi_{\theta_0}$
\FOR{step $= 1$ \TO $N_1$}
  \STATE Sample $\mathcal{G}=\{y^{(i)}\}_{i=1}^G$ for query $x$
  \FOR{each $y^{(i)}$}
    \STATE Compute $\{\entropy_t\}$ from rollout logits
    \STATE Compute $\ErrScore(y^{(i)})$ via Eq.~\eqref{eq:err_score}
    \STATE Assign $R_1(y^{(i)}, y^*)$ via Eq.~\eqref{eq:reward1}
  \ENDFOR
  \STATE Update $\pi_\theta$ via GRPO (Eq.~\eqref{eq:advantage})
\ENDFOR
\STATE \textbf{Phase 2 (\methodlen{}):}
\STATE $\pi_\theta \leftarrow \pi_{\theta_{N_1}}$
\FOR{step $= 1$ \TO $N_2$}
  \STATE Sample $\mathcal{G}$ for query $x$
  \STATE Compute $\mu_L$, $\sigma_L$ over $\{|y^{(i)}|\}_{i=1}^G$
  \FOR{each $y^{(i)}$}
    \STATE Compute $\lambda_i$ via
           Eqs.~\eqref{eq:zscore}--\eqref{eq:lambda}
    \STATE Assign $R_2(y^{(i)}, y^*)$ via Eq.~\eqref{eq:reward2}
  \ENDFOR
  \STATE Update $\pi_\theta$ via GRPO (Eq.~\eqref{eq:advantage})
\ENDFOR
\RETURN $\pi_\theta$
\end{algorithmic}
\end{algorithm}

\begin{table*}[t]
\centering
\small
\caption{Pass@1 accuracy (\%) and average response length (tokens,
\emph{Tok}) across model methods and five benchmarks.
\textbf{Bold} indicates the best accuracy per column;
\underline{underline} the shortest average length.}
\label{tab:main}
\setlength{\tabcolsep}{3.5pt}
\resizebox{\textwidth}{!}{%
\begin{tabular}{l cc cc cc cc cc cc}
\toprule
\multicolumn{1}{c}{\textbf{Method}} & \multicolumn{2}{c}{\textbf{GSM8K}}
& \multicolumn{2}{c}{\textbf{AIME24}}
& \multicolumn{2}{c}{\textbf{AMC23}}
& \multicolumn{2}{c}{\textbf{MATH-500}}
& \multicolumn{2}{c}{\textbf{MMLU-STEM}}
& \multicolumn{2}{c}{\textbf{Avg}} \\
\cmidrule(lr){2-3}\cmidrule(lr){4-5}\cmidrule(lr){6-7}
\cmidrule(lr){8-9}\cmidrule(lr){10-11}\cmidrule(lr){12-13}
 & Acc & Tok & Acc & Tok & Acc & Tok & Acc & Tok
& Acc & Tok & Acc & Tok \\
\midrule
\multicolumn{13}{c}{\cellcolor[rgb]{0.68, 0.85, 0.90}\textbf{DeepSeek-R1-Distill-Qwen-1.5B}} \\
\midrule
Base        & 84.6 & 2076 & 26.7 & 12958 & 70.0 & 8063 & 82.1 & 4729 & 51.7 & 6266 & 63.0 & 6818 \\
DEER        & 85.7 & 1553 & 26.7 & 10692 & 75.0 & \underline{5556} & 83.8 & 3534 & 50.4 & 6391 & 64.3 & 5545 \\
Dynasor     & 85.1 & 1916 & 26.7 & 11687 & 75.0 & 7287 & 82.0 & 4351 & 49.6 & 6670 & 63.7 & 6382 \\
PEAR        & 85.3 & 1949 & \textbf{33.3} & 12260 & 67.5 & 8015 & 81.0 & 4848 & 50.7 & 6033 & 63.6 & 6621 \\
GRPO        & 86.6 & 2281 & 30.0 & 12939 & 70.0 & 8336 & 82.6 & 5206 & 51.6 & 7440 & 64.2 & 7240 \\
GRPO+ERR+   & \textbf{88.6} & \underline{1324} & \textbf{33.3} & 10670 & \textbf{77.5} & 6288 & 84.8 & 3857 & \textbf{58.0} & 5109 & \textbf{68.4} & 5450 \\
Laser-L2048 & 82.8 & 1850 & 26.7 & \underline{10515} & 75.0 & 7503 & 80.3 & 4529 & 50.4 & 5632 & 63.0 & 6006 \\
DAPO        & 85.4 & 2123 & 26.7 & 12956 & 70.0 & 8065 & 83.4 & 4432 & 49.9 & 6966 & 63.1 & 6908 \\
DAPO+ERR+   & 87.1 & 1465 & 30.0 & 10542 & 75.0 & 5603 & \textbf{85.2} & \underline{3296} & 54.2 & \underline{4742} & 66.3 & \underline{5130} \\
\bottomrule
\end{tabular}%
}
\end{table*}
% =============================================================

% =============================================================
\section{Experiments}
\label{sec:experiments}

\subsection{Setup}
\label{sec:setup}

\paragraph{Models and Training.}
We evaluate across four widely adopted Large Reasoning Models:
DeepSeek-R1-Distill-Qwen-1.5B, DeepSeek-R1-Distill-Qwen-7B~\citep{deepseek-r1},
Qwen3-4B, and Qwen3-8B~\citep{qwen3}.
All models are trained using the open-source \texttt{verl}
framework on 7,473 problems from the GSM8K
training set~\citep{cobbe2021gsm8k}. We use a batch size of 128, a
learning rate of $1 \times 10^{-6}$, and a maximum response length of
16,384 tokens. For generation in the RL environment, we set
temperature to 0.6, top-$p$ to 0.95, rollouts per prompt to 8, and
KL penalty coefficient to 0.001. Default \method{} hyperparameters
are $\epsilon = 0.01$, $\gamma = 0.5$, $\lambda = 0.3$, $\alpha =
0.3$, $R_b = 1.0$, $R_f = 0.1$, and $R_{\max} = 1.5$ (phase-1 cap
only).
% We evaluate ERR+ across four backbones: DeepSeek-R1-Distill-Qwen-1.5B/7B \citep{deepseek-r1} and Qwen3-4B/8B~\citep{qwen3}. Training uses the open-source \texttt{verl} framework on 7,473 GSM8K training set~\citep{cobbe2021gsm8k}. Hyperparameters are: batch size 128, learning rate $1\times10^{-6}$, max length 16,384 tokens, temperature 0.6, top-$p$ 0.95, rollouts per prompt 8, and KL coefficient 0.001. Default ERR+ parameters are $\epsilon=0.01$, $\gamma=0.5$, $\lambda=0.3$, $\alpha=0.3$, $R_b=1.0$, $R_f=0.1$, and $R_{max}=1.5$ (capped in phase 1).
%============================================================
\paragraph{Evaluation Datasets.}
We benchmark on five datasets: (1)~\textbf{GSM8K}~\citep{cobbe2021gsm8k},
grade-school math word problems requiring 2--8 arithmetic steps;
(2)~\textbf{MATH-500}~\citep{hendrycks2021math}, 500 high-school
competition problems; (3)~\textbf{AIME24}~\citep{li2024numinamath},
30 problems from the 2024 American Invitational Mathematics
Examination; (4)~\textbf{AMC23}~\citep{li2024numinamath}, 40 problems
from the 2023 American Mathematics Competition; and
(5)~\textbf{MMLU-STEM}~\citep{hendrycks2021mmlu}, a STEM-focused
subset of MMLU for out-of-domain generalization evaluation. All
evaluations follow the Qwen2.5-Math evaluation
protocol~\citep{qwen25math} with temperature 0.6, top-$p$
0.95, and a length cap of 16,384 tokens. We report Pass@1 accuracy
(Acc) and average response length (Tok).

% \paragraph{Evaluation Datasets.}
% We benchmark on five datasets following the Qwen2.5-Math evaluation protocol~\citep{qwen25math} (temperature 0.6, top-$p$ 0.95, and a length cap of 16,384 tokens): (1)~\textbf{GSM8K}~\citep{cobbe2021gsm8k}, (2)~\textbf{MATH-500}~\citep{hendrycks2021math}, (3)~\textbf{AIME24}~\citep{li2024numinamath}, (4)~\textbf{AMC23}~\citep{li2024numinamath}, and (5)~\textbf{MMLU-STEM}~\citep{hendrycks2021mmlu} for out-of-domain generalization evaluation. We report Pass@1 accuracy (Acc) and average response length (Tok).

%------------------------------------
\paragraph{Baselines.}
We compare against: (1)~\textbf{Base}, the zero-shot model without
any RL training; (2)~\textbf{GRPO}~\citep{shao2024deepseekmath}, the
standard binary-reward RLVR baseline;
(3)~\textbf{PEAR}~\citep{huang2025pear}, which penalizes
thinking-phase entropy as a length proxy;
(4)~\textbf{DEER}~\citep{Yang2025DynamicEE}, \textbf{Dynasor}~\citep{Fu2024EfficientlySL}, and
\textbf{LASER-L2048}~\citep{luo2025laser}, representative
length-control baselines; and (5)~\textbf{DAPO}~\citep{yu2025dapo},
a strong entropy-aware RLVR baseline. We also include two ablation
variants: \emph{phase-2 only} (no phase-1 warm-up) and \emph{joint
training} ($R_1 + R_2$ from initialization).
All baselines and \method{} use the same length cap of $16{,}384$ tokens, temperature $0.6$, top-$p$ $0.95$, and the evaluation protocol in Section~\ref{sec:setup}, ensuring a fair comparison.

% ---------------------------------------------------------------

\subsection{Main Results}
\label{sec:main}
Table~\ref{tab:main} reports results on DeepSeek-R1-Distill-Qwen-1.5B against a broad set of
baselines. \method{} achieves highest accuracy while reducing length, showing the objectives need not trade off. Existing methods either improve accuracy without shortening responses (GRPO) or reduce length at the cost of accuracy (PEAR), whereas ERR+ simultaneously improves both metrics.
The gains transfer to DAPO as the base optimizer
(DAPO+\method{}), confirming that the reward signals are not specific
to GRPO.
Table~\ref{tab:main} and Appendix~E use the same model, data, and evaluation setup but differ only in aggregation: Table~\ref{tab:main} reports the best of four seeds (selected by validation accuracy with length monitored for outliers), while Appendix~E reports the four-run mean and standard deviation ($66.10\% \pm 0.61\%$, $5794 \pm 84$ tokens).

% \subsection{Main Results}
% \label{sec:main}

% Table~\ref{tab:main} shows that \method{} achieves the highest accuracy and shortest length on DeepSeek-R1-Distill-Qwen-1.5B, successfully avoiding the traditional accuracy-length trade-off seen in GRPO and PEAR. These gains transfer consistently to DAPO (DAPO+\method{}), confirming that our reward signals are framework-agnostic.

% ---------------------------------------------------------------
\subsection{Generalization Across Backbones}
\label{sec:backbone}

\begin{table*}[t]
\centering
\small
\caption{Pass@1 accuracy (\%) and average response length (tokens,
\emph{Tok}) across three model backbones and five benchmarks.
\textbf{Bold} indicates the best accuracy per backbone group;
\underline{underline} the shortest average length among trained
methods.}
\label{tab:main_other_models}
\setlength{\tabcolsep}{3.5pt}
\resizebox{\textwidth}{!}{%
\begin{tabular}{l cc cc cc cc cc cc}
\toprule
\multicolumn{1}{c}{\textbf{Method}} & \multicolumn{2}{c}{\textbf{GSM8K}}
& \multicolumn{2}{c}{\textbf{AIME24}}
& \multicolumn{2}{c}{\textbf{AMC23}}
& \multicolumn{2}{c}{\textbf{MATH-500}}
& \multicolumn{2}{c}{\textbf{MMLU-STEM}}
& \multicolumn{2}{c}{\textbf{Avg}} \\
\cmidrule(lr){2-3}\cmidrule(lr){4-5}\cmidrule(lr){6-7}
\cmidrule(lr){8-9}\cmidrule(lr){10-11}\cmidrule(lr){12-13}
 & Acc & Tok & Acc & Tok & Acc & Tok & Acc & Tok
& Acc & Tok & Acc & Tok \\
\midrule
\multicolumn{13}{c}{\cellcolor[rgb]{0.68, 0.85, 0.90}\textbf{Qwen3-4B}} \\
\midrule
Base      & 94.5 & 2078 & 56.7 & 11654 & 80.0 & 7704 & 89.8 & 4814 & 90.3 & 2471 & 82.3 & 5744 \\
GRPO      & \textbf{95.4} & 1969 & 53.3 & 11461 & 80.0 & 7615 & 90.2 & 4817 & 89.9 & 2573 & 81.8 & 5687 \\
PEAR      & 93.9 & \underline{1193} & 33.3 & \underline{10081} & \textbf{85.0} & \underline{5717} & 87.4 & \underline{3636} & \textbf{91.3} & \underline{1831} & 78.2 & \underline{4492} \\
GRPO+\method{} & 94.6 & 1898 & \textbf{60.0} & 11002 & \textbf{85.0} & 6648 & \textbf{91.0} & 4214 & 90.3 & 2205 & \textbf{84.2} & 4752 \\
\midrule
\multicolumn{13}{c}{\cellcolor[rgb]{0.68, 0.85, 0.90}\textbf{Qwen3-8B}} \\
\midrule
Base      & 93.9 & 2226 & 56.7 & 11877 & 80.0 & 8348 & 87.0 & 5088 & 92.6 & 2586 & 82.0 & 6025 \\
GRPO      & 94.6 & 2102 & 53.3 & 12576 & 80.5 & 8480 & 87.4 & 5631 & 92.6 & 2556 & 81.7 & 6269 \\
PEAR      & 93.8 & \underline{1639} & 36.7 & \underline{11579} & 77.5 & \underline{7281} & 85.6 & \underline{4388} & 92.4 & \underline{2206} & 77.2 & \underline{5418} \\
GRPO+\method{} & \textbf{96.0} & 1798 & \textbf{60.0} & 11619 & \textbf{82.5} & 7918 & \textbf{88.8} & 4675 & \textbf{92.9} & 2243 & \textbf{84.0} & 5451 \\
\midrule
\multicolumn{13}{c}{\cellcolor[rgb]{0.68, 0.85, 0.90}\textbf{DeepSeek-R1-Distill-Qwen-7B}} \\
\midrule
Base      & 90.3 & 1270 & 50.0 & 10185 & \textbf{92.5} & 5971 & 92.2 & 4092 & 70.8 & 3540 & 79.2 & 5012 \\
GRPO      & \textbf{93.6} & 1413 & 50.0 & 10934 & 90.0 & 7103 & 92.0 & 3915 & 71.2 & 4181 & 79.4 & 5509 \\
PEAR      & 91.5 & \underline{979} & 53.3 & 9903  & 90.0 & \underline{5214} & 91.6 & \underline{3213} & 73.8 & \underline{2489} & 80.0 & \underline{4360} \\
GRPO+\method{} & 92.7 & 1094 & \textbf{60.0} & \underline{9843}  & 90.0 & 5479 & \textbf{92.8} & 3289 & \textbf{77.0} & 2768 & \textbf{82.5} & 4495 \\
\bottomrule
\end{tabular}
}
\end{table*}

Table~\ref{tab:main_other_models} extends evaluation to three
additional backbones. \method{} achieves the highest average accuracy
on all three models while reducing response length compared to the
GRPO baseline. PEAR achieves the shortest responses on some backbones
but consistently degrades accuracy, particularly on harder benchmarks
such as AIME24. \method{} avoids this trade-off, confirming that the
two-phase design generalizes across model families and scales.

\begin{table*}[t]
\centering
\small
\caption{Ablation study of training phases on \texttt{DeepSeek-R1-Distill-Qwen-1.5B} across five datasets. Each row removes or modifies one design choice of \method{}. \textbf{Bold} indicates the best value per column.}
\label{tab:phase_ablation}
\setlength{\tabcolsep}{4.5pt}
\begin{tabular}{l cc cc cc cc cc cc}
\toprule
\multirow{2}{*}{\textbf{Variant}} & \multicolumn{2}{c}{\textbf{GSM8K}} & \multicolumn{2}{c}{\textbf{AIME24}} & \multicolumn{2}{c}{\textbf{AMC23}} & \multicolumn{2}{c}{\textbf{MATH-500}} & \multicolumn{2}{c}{\textbf{MMLU-STEM}} & \multicolumn{2}{c}{\textbf{Avg}} \\
\cmidrule(lr){2-3} \cmidrule(lr){4-5} \cmidrule(lr){6-7} \cmidrule(lr){8-9} \cmidrule(lr){10-11} \cmidrule(lr){12-13}
 & \textbf{Acc} & \textbf{Tok} & \textbf{Acc} & \textbf{Tok} & \textbf{Acc} & \textbf{Tok} & \textbf{Acc} & \textbf{Tok} & \textbf{Acc} & \textbf{Tok} & \textbf{Acc} & \textbf{Tok} \\
\midrule
\method{} (full)             & \textbf{88.6} & 1324          & \textbf{33.3} & \textbf{10670} & \textbf{77.5} & \textbf{6288} & \textbf{84.8} & \textbf{3857} & \textbf{58.0} & 5109          & \textbf{68.4} & \textbf{5450} \\
\midrule
w/o phase-1 (phase-2 only)   & 85.0          & \textbf{1244} & 26.7          & 10729          & 67.5          & 6933          & 84.0          & 4253          & 51.4          & \textbf{4168} & 62.9          & 5465          \\
w/o phase-2 (phase-1 only)   & 88.1          & 1612          & \textbf{33.3} & 12807          & 72.5          & 8061          & 84.6          & 4752          & 55.6          & 5895          & 66.8          & 6625          \\
Joint training ($R_1 + R_2$) & 84.2          & 2005          & 23.3          & 12105          & 72.5          & 7245          & 83.0          & 4470          & 49.9          & 6470          & 62.6          & 6459          \\
\bottomrule
\end{tabular}
\end{table*}
%=======================================
\subsection{Ablation Study}
\label{sec:ablation}

Table~\ref{tab:phase_ablation} presents results validating our design choices on the 1.5B backbone.
\textbf{Phase-1 Necessity:} Removing phase-1 (phase-2 only) achieves the shortest response length on GSM8K (1244 tokens, 6\% shorter than full \method{}), but its average length across all benchmarks (5465 tokens) is comparable to full \method{} (5450 tokens) and suffers significant accuracy degradation: 5.5\%
average accuracy loss compared to the full model. This confirms that premature length compression without a stable reasoning structure sacrifices quality for brevity.
\textbf{Phase-2 Necessity:} Omitting phase-2 (phase-1 only) maintains strong accuracy (66.8\%) but fails to compress responses, resulting in 21.5\% longer traces than full \method{}.
\textbf{Sequential vs. Joint Training:} Joint training ($R_1 + R_2$ simultaneously) performs worst overall, showing 5.8\% lower accuracy and 18.5\% longer responses than full \method{}, which validates Theorem 1's prediction of early gradient conflict.
\textbf{Summary:} The ablation study confirms that both phases are
necessary: phase-1 establishes correct, decisive reasoning via entropy
drops; phase-2 refines length via group-relative penalties. The
sequential design is substantially superior to any individual component
or joint optimization.

\section{Ablation on Length Normalization in ERR}
\label{app:norm}

The \methodbase{} score (Eq.~\eqref{eq:err_score}) uses
$\log(T_k+1)$ as a length normalizer.
We consider two failure modes: (i)~no normalization rewards
long traces for accumulating drops regardless of quality;
(ii)~linear normalization $(\sum r_t)/T_k$ imposes a
constant per-token drop-density requirement and unfairly
penalizes hard problems that genuinely need longer exploration.
Logarithmic normalization provides a middle ground:
the per-token density requirement diminishes as traces
lengthen, rewarding milestone-like uncertainty resolution
rather than verbosity, without enforcing a fixed density target.

\begin{table}[h]
\centering
\footnotesize
\caption{Effect of ERR length normalizer on average accuracy
  and token count across five benchmarks
  (\texttt{DeepSeek-R1-Distill-Qwen-1.5B}).}
\label{tab:norm}
\begin{tabularx}{\columnwidth}{@{}l Y Y@{}}
\toprule
\textbf{Normalizer} & \textbf{Avg Acc (\%)} & \textbf{Avg Tok} \\
\midrule
$T_k$ (linear)    & 48.6 & 2063 \\
$\sqrt{T_k}$      & 53.3 & 3314 \\
$\log(T_k+1)$     & \textbf{68.4} & \textbf{5450} \\
\bottomrule
\end{tabularx}
\end{table}

Table~\ref{tab:norm} compares the three choices on
\texttt{DeepSeek-R1-Distill-Qwen-1.5B}. Linear normalization imposes the strictest per-token density
requirement, which over-penalizes hard problems and causes
significant accuracy degradation.
$\sqrt{T_k}$ alleviates this but still shows clear accuracy
loss relative to the logarithmic form.
$\log(T_k+1)$ gives the best trade-off, allowing phase~1
to build correct reasoning structure while phase~2 handles
subsequent length compression.
This ablation isolates the \emph{Phase-1 normalizer}: the $\log(T_k+1)$
form applies a milder sub-linear penalty, giving genuinely complex
multi-step reasoning a grace region and avoiding harmful
over-compression, whereas linear normalization yields shorter outputs
only at a large accuracy cost. The final response efficiency of
\method{} is achieved by Phase-2 RRER \emph{after} Phase~1 has
established a stable reasoning structure, not by the Phase-1 normalizer
alone.

\begin{table}[h]
\centering
\small
\caption{Pass@1 accuracy (\%) and average response length
(tokens, \emph{Tok}) on MMLU-STEM and GPQA Diamond.
\textbf{Bold} indicates the best accuracy;
\label{tab:scope_generality}
\underline{underline} the shortest average length.}
\setlength{\tabcolsep}{8pt}
\begin{tabular}{l cc cc}
\toprule
\multicolumn{1}{c}{\textbf{Method}}
  & \multicolumn{2}{c}{\textbf{MMLU-STEM}}
  & \multicolumn{2}{c}{\textbf{GPQA Diamond}} \\
\cmidrule(lr){2-3}\cmidrule(lr){4-5}
 & Acc & Tok & Acc & Tok \\
\midrule
\multicolumn{5}{c}{\cellcolor[rgb]{0.68, 0.85, 0.90}%
  \textbf{DeepSeek-R1-Distill-Qwen-1.5B}} \\
\midrule
Base       & 51.7 & 6266  & 27.8 & 13875 \\
GRPO       & 51.6 & 7440  & 29.3 & 14528 \\
GRPO+\method{} & \textbf{58.0} & \underline{5109}
           & \textbf{31.6} & \underline{10961} \\
\bottomrule
\end{tabular}
\end{table}

\subsection{Anti-Reward-Hacking Analysis}
\label{sec:antihack}

A potential concern is that ERR optimizes entropy drops as a proxy signal rather than reasoning quality directly, creating an incentive for the policy to generate
artificially sharp entropy transitions (e.g.\ inflate-then-drop cycles, repetitive
deterministic patterns) without genuinely improving reasoning.
We address this with an intervention check: if such exploitation occurred, the ratio
of upward entropy movement to rewarded drops (\emph{Up/Drop}) and the per-trace sign-change
count should inflate without corresponding accuracy gains.
Table~\ref{tab:antihack} shows that neither occurs.
% To check this, an abnormal inflation in the upward-to-rewarded entropy ratio (\emph{Up/Drop}) or sign-change counts would signal hacking; Table~\ref{tab:antihack} shows neither occurs.
Up/Drop remains $\approx 1.0$ throughout training and sign changes decrease steadily,
indicating that rewarded drops track genuine commitment events rather than
artificial cycling. This is consistent with recent evidence that high-entropy
forking tokens carry disproportionate functional impact on reasoning
outcomes~\citep{wang2025beyond,chen2025unreasonable}: ERR rewards their resolution in
correct traces only, without penalizing exploration or directly minimizing entropy.

\begin{table}[t]
\centering
\small  % 或 \footnotesize，根据空间微调
\caption{Anti-reward-hacking check on \textsc{GSM8K}
  (\texttt{DeepSeek-R1-Distill-Qwen-1.5B}).
  \emph{Up} and \emph{Drop} are the per-trace totals of upward entropy movement
  and rewarded entropy drops, respectively.
  \emph{Sign chg.}\ counts entropy-direction switches per trace.
  If ERR exploited artificial cycles, Up/Drop and Sign chg.\ should rise;
  both decrease instead.}
\label{tab:antihack}
\setlength{\tabcolsep}{0pt}  % 关键：先清零，让 \extracolsep 接管间距
\begin{tabular*}{\columnwidth}{@{\extracolsep{\fill}}lcccccc@{}}
\toprule
\textbf{Method} & \textbf{Acc} & \textbf{Tok}
  & \textbf{Up} & \textbf{Drop} & \textbf{Up/Drop} & \textbf{Sign chg.} \\
\midrule
GRPO              & 86.6 & 2281 & 507.2 & 506.9 & 1.0007 & 1012.98 \\
ERR (Phase~1)     & 88.1 & 1612 & 281.0 & 280.8 & 1.0010 & 683.06  \\
\method{}         & 88.6 & 1324 & 250.2 & 250.0 & 1.0011 & 609.77  \\
\bottomrule
\end{tabular*}
\end{table}

% ============================================================
\section{Conclusion}
\label{sec:conclusion}

We presented \method{}, a two-phase RLVR framework for large reasoning
models. The method is grounded in the observation that correct
reasoning traces exhibit more frequent and larger token-level entropy
drops than incorrect ones. Phase~1 rewards entropy resolution in the
thinking phase without constraining high-entropy exploration. Phase~2 introduces a difficulty-aware, outlier-robust
length signal by scoring response length against co-generated peers.
A formal gradient-conflict analysis establishes that sequential
application is preferable to joint optimization in early training, and
ablations confirm both empirical predictions. Experiments on
GSM8K, AIME~2024, AMC23, MATH-500, and MMLU-STEM show consistent
accuracy improvements over binary-reward and entropy-based baselines
alongside meaningful length reductions.

\section*{Limitations}

While \method{} demonstrates consistent improvements in accuracy and conciseness across multiple reasoning benchmarks, several limitations warrant discussion and point to promising avenues for future work. Our optimization objectives target accuracy and response length, but human evaluators may also value attributes such as step-by-step explainability, pedagogical clarity, or stylistic consistency. Integrating multi-objective rewards that balance efficiency with human-centric qualities, potentially via preference learning or multi-task reinforcement learning, represents an important direction for deploying reasoning models in real-world applications.

\section*{Acknowledgments}

This work was supported by New Generation Artificial Intelligence–National Science and Technology Major Project (2025ZD0124103) in collaboration with Shanghai Artificial Intelligence Laboratory. It is also supported by National Natural Science Foundation of China (under Projects No.~62477011 and 62377013), and the Fundamental Research Funds for the Central Universities.

\vspace{10em}

\bibliography{references}

\begin{thebibliography}{29}
\providecommand{\natexlab}[1]{#1}

\bibitem[{Agarwal et~al.(2025)Agarwal, Zhang, Yuan, Han, and Peng}]{chen2025unreasonable}
Shivam Agarwal, Zimin Zhang, Lifan Yuan, Jiawei Han, and Hao Peng. 2025.
\newblock The unreasonable effectiveness of entropy minimization in {LLM} reasoning.
\newblock \emph{Advances in Neural Information Processing Systems}, 38.

\bibitem[{Cheng et~al.(2025)Cheng, Huang, Zhu, Dai, Zhao, Zhang, and Wei}]{cheng2025reasoning}
Daixuan Cheng, Shaohan Huang, Xuekai Zhu, Bo~Dai, Wayne~Xin Zhao, Zhenliang Zhang, and Furu Wei. 2025.
\newblock Reasoning with exploration: An entropy perspective.
\newblock \emph{arXiv preprint arXiv:2506.14758}.

\bibitem[{Cobbe et~al.(2021)Cobbe, Kosaraju, Bavarian, Chen, Jun, Kaiser, Plappert, Tworek, Hilton, Nakano, Hesse, and Schulman}]{cobbe2021gsm8k}
Karl Cobbe, Vineet Kosaraju, Mo~Bavarian, Mark Chen, Heewoo Jun, Lukasz Kaiser, Matthias Plappert, Jerry Tworek, Jacob Hilton, Reiichiro Nakano, Christopher Hesse, and John Schulman. 2021.
\newblock \href {https://api.semanticscholar.org/CorpusID:239998651} {Training verifiers to solve math word problems}.
\newblock \emph{ArXiv}, abs/2110.14168.

\bibitem[{Cui et~al.(2025)Cui, Zhang, Chen, Yuan, Wang, Zuo, Li, Fan, Chen, Chen, Liu, Peng, Bai, Ouyang, Cheng, Zhou, and Ding}]{cui2025entropy}
Ganqu Cui, Yuchen Zhang, Jiacheng Chen, Lifan Yuan, Zhi Wang, Yuxin Zuo, Haozhan Li, Yuchen Fan, Huayu Chen, Weize Chen, Zhiyuan Liu, Hao Peng, Lei Bai, Wanli Ouyang, Yu~Cheng, Bowen Zhou, and Ning Ding. 2025.
\newblock \href {https://arxiv.org/abs/2505.22617} {The entropy mechanism of reinforcement learning for reasoning language models}.
\newblock \emph{Preprint}, arXiv:2505.22617.

\bibitem[{{DeepSeek-AI}(2025)}]{deepseek-r1}
{DeepSeek-AI}. 2025.
\newblock {DeepSeek-R1}: Incentivizing reasoning capability in {LLMs} via reinforcement learning.
\newblock \emph{arXiv preprint arXiv:2501.12948}.

\bibitem[{Fu et~al.(2024)Fu, Chen, Zhu, Fu, Dai, Zhuang, Ma, Qiao, Rosing, Stoica, and Zhang}]{Fu2024EfficientlySL}
Yichao Fu, Junda Chen, Siqi Zhu, Zheyu Fu, Zhongdongming Dai, Yonghao Zhuang, Yi~Ma, Aurick Qiao, Tajana Rosing, Ion Stoica, and Hao Zhang. 2024.
\newblock \href {https://api.semanticscholar.org/CorpusID:275133390} {Efficiently scaling llm reasoning with certaindex}.

\bibitem[{Hendrycks et~al.(2020)Hendrycks, Burns, Basart, Zou, Mazeika, Song, and Steinhardt}]{hendrycks2021mmlu}
Dan Hendrycks, Collin Burns, Steven Basart, Andy Zou, Mantas Mazeika, Dawn~Xiaodong Song, and Jacob Steinhardt. 2020.
\newblock \href {https://api.semanticscholar.org/CorpusID:221516475} {Measuring massive multitask language understanding}.
\newblock \emph{ArXiv}, abs/2009.03300.

\bibitem[{Hendrycks et~al.(2021)Hendrycks, Burns, Kadavath, Arora, Basart, Tang, Song, and Steinhardt}]{hendrycks2021math}
Dan Hendrycks, Collin Burns, Saurav Kadavath, Akul Arora, Steven Basart, Eric Tang, Dawn~Xiaodong Song, and Jacob Steinhardt. 2021.
\newblock \href {https://api.semanticscholar.org/CorpusID:232134851} {Measuring mathematical problem solving with the math dataset}.
\newblock \emph{ArXiv}, abs/2103.03874.

\bibitem[{Hu et~al.(2026)Hu, Qiu, Xu, Li, Zhou, and King}]{hu2025conmax}
Minda Hu, Zexuan Qiu, Zenan Xu, Kun Li, Bo~Zhou, and Irwin King. 2026.
\newblock \href {https://api.semanticscholar.org/CorpusID:284544288} {Conmax: Confidence-maximizing compression for efficient chain-of-thought reasoning}.
\newblock \emph{ArXiv}, abs/2601.04973.

\bibitem[{Huang et~al.(2025)Huang, Lu, and Zhang}]{huang2025pear}
Chen Huang, Wei Lu, and Wenxuan Zhang. 2025.
\newblock {PEAR}: Phase entropy aware reward for efficient reasoning.
\newblock \emph{arXiv preprint arXiv:2510.08026}.

\bibitem[{Li et~al.()Li, Beeching, Tunstall, Lipkin, Soletskyi, Huang, Rasul, Yu, Jiang, Shen, Qin, Dong, Zhou, Fleureau, Lample, Polu, Face, and Mistral}]{li2024numinamath}
Jia Li, Edward Beeching, Lewis Tunstall, Ben Lipkin, Roman Soletskyi, Shengyi Huang, Kashif Rasul, Long Yu, Albert~Qiaochu Jiang, Ziju Shen, Zihan Qin, Bin Dong, Li~Zhou, Yann Fleureau, Guillaume Lample, Stanislas Polu, Hugging Face, and AI~Mistral.
\newblock \href {https://api.semanticscholar.org/CorpusID:286088415} {Numinamath: The largest public dataset in ai4maths with 860k pairs of competition math problems and solutions}.

\bibitem[{Liu et~al.(2026{\natexlab{a}})Liu, Dong, Lu, Diao, Belc{\'a}k, Liu, Chen, Yin, Wang, Cheng, Choi, Kautz, and Molchanov}]{liu2025gdpo}
Shih-Yang Liu, Xin Dong, Ximing Lu, Shizhe Diao, Peter Belc{\'a}k, Mingjie Liu, Min-Hung Chen, Hongxu Yin, Yu-Chiang~Frank Wang, Kwang-Ting Cheng, Yejin Choi, Jan Kautz, and Pavlo Molchanov. 2026{\natexlab{a}}.
\newblock \href {https://api.semanticscholar.org/CorpusID:284543897} {Gdpo: Group reward-decoupled normalization policy optimization for multi-reward rl optimization}.
\newblock \emph{ArXiv}, abs/2601.05242.

\bibitem[{Liu et~al.(2026{\natexlab{b}})Liu, Zhou, Deng, Huang, Liu, Deng, Zhang, and He}]{luo2025laser}
Wei Liu, Ruochen Zhou, Yiyun Deng, Yuzhen Huang, Junteng Liu, Yuntian Deng, Yizhe Zhang, and Junxian He. 2026{\natexlab{b}}.
\newblock Learn to reason efficiently with adaptive length-based reward shaping.
\newblock In \emph{International Conference on Learning Representations}, volume 2026, pages 43279--43308.

\bibitem[{Liu et~al.(2025)Liu, Chen, Li, Qi, Pang, Du, Lee, and Lin}]{liu2025drgrpo}
Zichen Liu, Changyu Chen, Wenjun Li, Penghui Qi, Tianyu Pang, Chao Du, Wee~Sun Lee, and Min Lin. 2025.
\newblock Understanding r1-zero-like training: A critical perspective.
\newblock \emph{arXiv preprint arXiv:2503.20783}.

\bibitem[{{OpenAI}(2024)}]{openai-o1}
{OpenAI}. 2024.
\newblock \href {https://openai.com/index/learning-to-reason-with-llms/} {Learning to reason with {LLMs}}.
\newblock \emph{OpenAI Blog}.

\bibitem[{Rein et~al.(2023)Rein, Hou, Stickland, Petty, Pang, Dirani, Michael, and Bowman}]{Rein2023GPQAAG}
David Rein, Betty~Li Hou, Asa~Cooper Stickland, Jackson Petty, Richard~Yuanzhe Pang, Julien Dirani, Julian Michael, and Samuel~R. Bowman. 2023.
\newblock \href {https://api.semanticscholar.org/CorpusID:265295009} {Gpqa: A graduate-level google-proof q\&a benchmark}.
\newblock \emph{ArXiv}, abs/2311.12022.

\bibitem[{Shao et~al.(2024)Shao, Wang, Zhu, Xu, Song, Zhang, Li, Wu, and Guo}]{shao2024deepseekmath}
Zhihong Shao, Peiyi Wang, Qihao Zhu, Runxin Xu, Jun-Mei Song, Mingchuan Zhang, Y.~K. Li, Yu~Wu, and Daya Guo. 2024.
\newblock \href {https://api.semanticscholar.org/CorpusID:267412607} {Deepseekmath: Pushing the limits of mathematical reasoning in open language models}.
\newblock \emph{ArXiv}, abs/2402.03300.

\bibitem[{Tan and Pan(2025)}]{tan2025gtpo}
Hongze Tan and Jianfei Pan. 2025.
\newblock {GTPO} and {GRPO-S}: Token and sequence-level reward shaping with policy entropy.
\newblock \emph{arXiv preprint arXiv:2508.04349}.

\bibitem[{Team et~al.(2025)Team, Du, Gao, Xing, Jiang, Chen, Li, Xiao, Du, Liao, Tang, Wang, Zhang, Yuan, Lu, Tang, Sung, Wei, Lai, Guo, Zhu, Ding, Hu, Yang, Zhang, Yao, Zhao, Lu, Li, Yu, Gao, Zheng, Yuan, Chen, Guo, Su, Wang, Zhao, Zhang, Liu, Yan, Wu, Shi, Ye, Yu, Dong, Zhang, Ma, Pan, Gong, Liu, Ma, Wei, Cao, Huang, Jiang, Gao, Xiong, He, Huang, Xu, Wu, He, Wei, Jia, Wu, Xu, Zu, Zhou, Pan, Charles, Li, Hu, Liu, Chen, Wang, Liu, Qin, Liu, Yang, Bao, Du, Wu, Wang, Zhou, Wang, Li, Zhu, Zhang, Wang, Yang, Huang, Huang, Xu, Yang, and Lin}]{kimi15}
Kimi Team, Angang Du, Bofei Gao, Bowei Xing, Changjiu Jiang, Cheng Chen, Cheng Li, Chenjun Xiao, Chenzhuang Du, Chonghua Liao, Chuning Tang, Congcong Wang, Dehao Zhang, Enming Yuan, Enzhe Lu, Fengxiang Tang, Flood Sung, Guangda Wei, Guokun Lai, and 77 others. 2025.
\newblock \href {https://arxiv.org/abs/2501.12599} {Kimi k1.5: Scaling reinforcement learning with llms}.
\newblock \emph{Preprint}, arXiv:2501.12599.

\bibitem[{Wang et~al.(2025)Wang, Yu, Gao, Zheng, Liu, Lu, Dang, Chen, Yang, Zhang, Liu, Yang, Zhao, Yue, Song, Yu, Huang, and Lin}]{wang2025beyond}
Shenzhi Wang, Le~Yu, Chang Gao, Chujie Zheng, Shixuan Liu, Rui Lu, Kai Dang, Xiong-Hui Chen, Jianxin Yang, Zhenru Zhang, Yuqiong Liu, An~Yang, Andrew Zhao, Yang Yue, Shiji Song, Bowen Yu, Gao Huang, and Junyang Lin. 2025.
\newblock Beyond the 80/20 rule: High-entropy minority tokens drive effective reinforcement learning for {LLM} reasoning.
\newblock \emph{Advances in Neural Information Processing Systems}, 38.

\bibitem[{Wei et~al.(2022)Wei, Wang, Schuurmans, Bosma, Chi, Xia, Le, and Zhou}]{wei2022chain}
Jason Wei, Xuezhi Wang, Dale Schuurmans, Maarten Bosma, Ed~H. Chi, F.~Xia, Quoc Le, and Denny Zhou. 2022.
\newblock \href {https://api.semanticscholar.org/CorpusID:246411621} {Chain of thought prompting elicits reasoning in large language models}.
\newblock \emph{ArXiv}, abs/2201.11903.

\bibitem[{Yang et~al.(2025{\natexlab{a}})Yang, Li, Yang, Zhang, Hui, Zheng, Yu, Gao, Huang, Lv, Zheng, Liu, Zhou, Huang, Hu, Ge, Wei, Lin, Tang, Yang, Tu, Zhang, Yang, Yang, Zhou, Zhou, Lin, Dang, Bao, Yang, Yu, Deng, Li, Xue, Li, Zhang, Wang, Zhu, Men, Gao, Liu, Luo, Li, Tang, Yin, Ren, Wang, Zhang, Ren, Fan, Su, Zhang, Zhang, Wan, Liu, Wang, Cui, Zhang, Zhou, and Qiu}]{qwen3}
An~Yang, Anfeng Li, Baosong Yang, Beichen Zhang, Binyuan Hui, Bo~Zheng, Bowen Yu, Chang Gao, Chengen Huang, Chenxu Lv, Chujie Zheng, Dayiheng Liu, Fan Zhou, Fei Huang, Feng Hu, Hao Ge, Haoran Wei, Huan Lin, Jialong Tang, and 41 others. 2025{\natexlab{a}}.
\newblock \href {https://arxiv.org/abs/2505.09388} {Qwen3 technical report}.
\newblock \emph{Preprint}, arXiv:2505.09388.

\bibitem[{Yang et~al.(2024)Yang, Zhang, Hui, Gao, Yu, Li, Liu, Tu, Zhou, Lin, Lu, Xue, Lin, Liu, Ren, and Zhang}]{qwen25math}
An~Yang, Beichen Zhang, Binyuan Hui, Bofei Gao, Bowen Yu, Chengpeng Li, Dayiheng Liu, Jianhong Tu, Jingren Zhou, Junyang Lin, Keming Lu, Mingfeng Xue, Runji Lin, Tianyu Liu, Xingzhang Ren, and Zhenru Zhang. 2024.
\newblock \href {https://arxiv.org/abs/2409.12122} {Qwen2.5-math technical report: Toward mathematical expert model via self-improvement}.
\newblock \emph{Preprint}, arXiv:2409.12122.

\bibitem[{Yang et~al.(2025{\natexlab{b}})Yang, Si, Duan, Zhu, Zhu, Lin, Cao, and Wang}]{Yang2025DynamicEE}
Chenxu Yang, Qingyi Si, Yongjie Duan, Zheliang Zhu, Chenyu Zhu, Zheng Lin, Li~Cao, and Weiping Wang. 2025{\natexlab{b}}.
\newblock \href {https://api.semanticscholar.org/CorpusID:277994255} {Dynamic early exit in reasoning models}.
\newblock \emph{ArXiv}, abs/2504.15895.

\bibitem[{Yu et~al.(2025)Yu, Zhang, Zhu, Yuan, Zuo, Yue, Dai, Fan, Liu, Liu, Liu, Lin, Lin, Ma, Sheng, Tong, Zhang, Zhang, Zhang, Zhu, Zhu, Chen, Chen, Wang, Yu, Song, Wei, Zhou, Liu, Ma, Zhang, Yan, Qiao, Wu, and Wang}]{yu2025dapo}
Qiying Yu, Zheng Zhang, Ruofei Zhu, Yufeng Yuan, Xiaochen Zuo, Yu~Yue, Weinan Dai, Tiantian Fan, Gaohong Liu, Lingjun Liu, Xin Liu, Haibin Lin, Zhiqi Lin, Bole Ma, Guangming Sheng, Yuxuan Tong, Chi Zhang, Mofan Zhang, Wang Zhang, and 16 others. 2025.
\newblock \href {https://arxiv.org/abs/2503.14476} {Dapo: An open-source llm reinforcement learning system at scale}.
\newblock \emph{Preprint}, arXiv:2503.14476.

\bibitem[{Yue et~al.(2025)Yue, Yuan, Yu, Zuo, Zhu, Xu, Chen, Wang, Fan, Du, Wei, Yu, Liu, Liu, Liu, Lin, Lin, Ma, Zhang, Zhang, Zhang, Zhu, Zhang, Liu, Wang, Wu, and Yan}]{yu2025vapo}
Yu~Yue, Yufeng Yuan, Qiying Yu, Xiaochen Zuo, Ruofei Zhu, Wenyuan Xu, Jiaze Chen, Chengyi Wang, TianTian Fan, Zhengyin Du, Xiangpeng Wei, Xiangyu Yu, Gaohong Liu, Juncai Liu, Lingjun Liu, Haibin Lin, Zhiqi Lin, Bole Ma, Chi Zhang, and 8 others. 2025.
\newblock \href {https://arxiv.org/abs/2504.05118} {Vapo: Efficient and reliable reinforcement learning for advanced reasoning tasks}.
\newblock \emph{Preprint}, arXiv:2504.05118.

\bibitem[{Zhang et~al.(2025{\natexlab{a}})Zhang, Wen, Wu, and Huang}]{zhang2025edge}
Xingjian Zhang, Siwei Wen, Wenjun Wu, and Lei Huang. 2025{\natexlab{a}}.
\newblock {Edge-GRPO}: Entropy-driven {GRPO} with guided error correction for advantage diversity.
\newblock \emph{arXiv preprint arXiv:2507.21848}.

\bibitem[{Zhang et~al.(2025{\natexlab{b}})Zhang, Zhang, Guan, Cheng, Duan, Wang, Wang, Zheng, and He}]{zhang2025nofree}
Yanzhi Zhang, Zhaoxi Zhang, Haoxiang Guan, Yilin Cheng, Yitong Duan, Chen Wang, Yue Wang, Shuxin Zheng, and Jiyan He. 2025{\natexlab{b}}.
\newblock No free lunch: Rethinking internal feedback for {LLM} reasoning.
\newblock \emph{arXiv preprint arXiv:2506.17219}.

\bibitem[{Zhang et~al.(2025{\natexlab{c}})Zhang, Yu, Pan, Jin, Fu, Cai, Lin, and Ye}]{tokensqueeze2025}
Yuxiang Zhang, Zhengxu Yu, Weihang Pan, Zhongming Jin, Qiang Fu, Deng Cai, Binbin Lin, and Jieping Ye. 2025{\natexlab{c}}.
\newblock \href {https://api.semanticscholar.org/CorpusID:283073440} {Tokensqueeze: Performance-preserving compression for reasoning llms}.
\newblock \emph{ArXiv}, abs/2511.13223.

\end{thebibliography}

% ============================================================
\appendix

\section{Proof of Theorem~\ref{thm:conflict}}
\label{app:proof}

All notation follows Section~\ref{sec:theory}. We write
$F(\theta) \triangleq \mathbb{E}_{y \sim \pi_\theta}[\ErrScore(y)]$,
so that $\mathcal{L}_1(\theta) = -\mathbb{E}[R_1] = -R_b\,p_c -
\lambda F(\theta)$, where $p_c$ is the correctness probability
(treated as approximately constant for small $\eta$). Hence
$g_1 = \nabla_\theta \mathcal{L}_1 = -\lambda\,\nabla_\theta F(\theta)$.

Since $R_2 \propto \tanh(-\gamma z_i)$ is strictly decreasing in
$L_i = |\mathcal{T}(y)| + (T - T_k)$, we have
$\nabla_{L_i}\mathbb{E}[R_2] < 0$, so $g_2$ assigns negative
log-probability weight to tokens that increase $L_i$. Among tokens
in $\mathcal{T}(y)$, those in $\mathcal{T}_\rho(y)$ satisfy $r_t = 0$
and thus contribute nothing to $R_1$; they bear the highest
ratio of length cost to reward benefit and are therefore the primary
candidates for suppression under $g_2$.

Consider the update $\theta' = \theta - \eta g_2$. For each
$t \in \mathcal{T}_\rho(y)$, let $t'$ be the next committed position
satisfying $r_{t'} > 0$. Under the autoregressive factorization,
$\entropy_{t'-1}$ is a function of the generation context up to
position $t'-1$, which depends on the token at $t$. Reducing the
probability mass on position $t$ under $g_2$ decreases
$\mathbb{E}[\entropy_{t'-1}]$ by at least $\Delta\bar{\entropy}$ in
expectation, and hence decreases
$r_{t'} = \max(\entropy_{t'-1} - \entropy_{t'} - \epsilon, 0)$
by at least $\Delta\bar{\entropy}$. Summing over the $\rho(\theta)$
fraction of such positions and dividing by $\log(T_k+1)$:
\begin{align}
  \small
  F(\theta') \;\leq\; F(\theta)
  - \frac{\eta\,\rho(\theta)\,\Delta\bar{\entropy}}{\log(T_k+1)}.
  \label{eq:err_drop}
\end{align}

Taking $\eta \to 0$ in Eq.~\eqref{eq:err_drop}:
\begin{equation}
  \small
  \nabla_\theta F(\theta) \cdot (-g_2)
  \;\leq\; -\frac{\rho(\theta)\,\Delta\bar{\entropy}}{\log(T_k+1)}.
\end{equation}
Since $g_1 = -\lambda\,\nabla_\theta F$, we have
$\nabla_\theta F = -g_1/\lambda$, so:
\begin{equation}
  \small
  \frac{g_1}{\lambda} \cdot g_2
  \;\leq\; -\frac{\rho(\theta)\,\Delta\bar{\entropy}}{\log(T_k+1)},
\end{equation}
which gives Eq.~\eqref{eq:conflict} with
$C = \lambda\,\Delta\bar{\entropy}/\log(T_k+1) > 0$.

By a first-order Taylor expansion:
\begin{align}
  \small
  &\mathcal{L}_1\!\left(\theta - \eta(g_1 + g_2)\right) \notag\\
  &\quad= \mathcal{L}_1(\theta)
     - \eta\langle g_1,\, g_1 + g_2 \rangle + O(\eta^2) \notag\\
  &\quad= \mathcal{L}_1(\theta) - \eta\|g_1\|^2
     - \eta\langle g_1, g_2\rangle + O(\eta^2) \notag\\
  &\quad\geq \mathcal{L}_1(\theta) - \eta\|g_1\|^2
     + \eta C\rho(\theta) + O(\eta^2).
  \label{eq:taylor}
\end{align}
The single-objective step satisfies
$\mathcal{L}_1(\theta - \eta g_1) = \mathcal{L}_1(\theta) -
\eta\|g_1\|^2 + O(\eta^2)$, so subtracting the two expressions
yields Eq.~\eqref{eq:degradation}.

For the convergence claim: at $\theta_1^*$, every high-entropy
position in a correct trace is followed by a committed drop, so
$\mathcal{T}_\rho(y) = \varnothing$ and $\rho(\theta_1^*) = 0$.
The right-hand sides of Eqs.~\eqref{eq:conflict}
and~\eqref{eq:degradation} both vanish. Furthermore, the tokens
that $g_2$ would remove at $\theta_1^*$ are low-entropy, low-$r_t$
spans between commitment points; by Eq.~\eqref{eq:err_drop} with
$\rho = 0$, these removals leave $F(\theta_1^*)$ unchanged, so
applying $R_2$ from $\theta_1^*$ incurs no degradation to
$\mathcal{L}_1$. \hfill$\square$

\begin{remark}
$C = \lambda\,\Delta\bar{\entropy}/\log(T_k+1)$ grows with the
entropy surplus $\Delta\bar{\entropy}$ and decreases with thinking
length $T_k$. Models with more volatile exploration (larger
$\Delta\bar{\entropy}$) or shorter thinking budgets benefit most
from the sequential schedule.
\end{remark}

\section{Hyperparameter Sensitivity Analysis}
\label{sec:hyperparam}

\begin{figure*}[t]
  \centering
  \includegraphics[width=\textwidth]{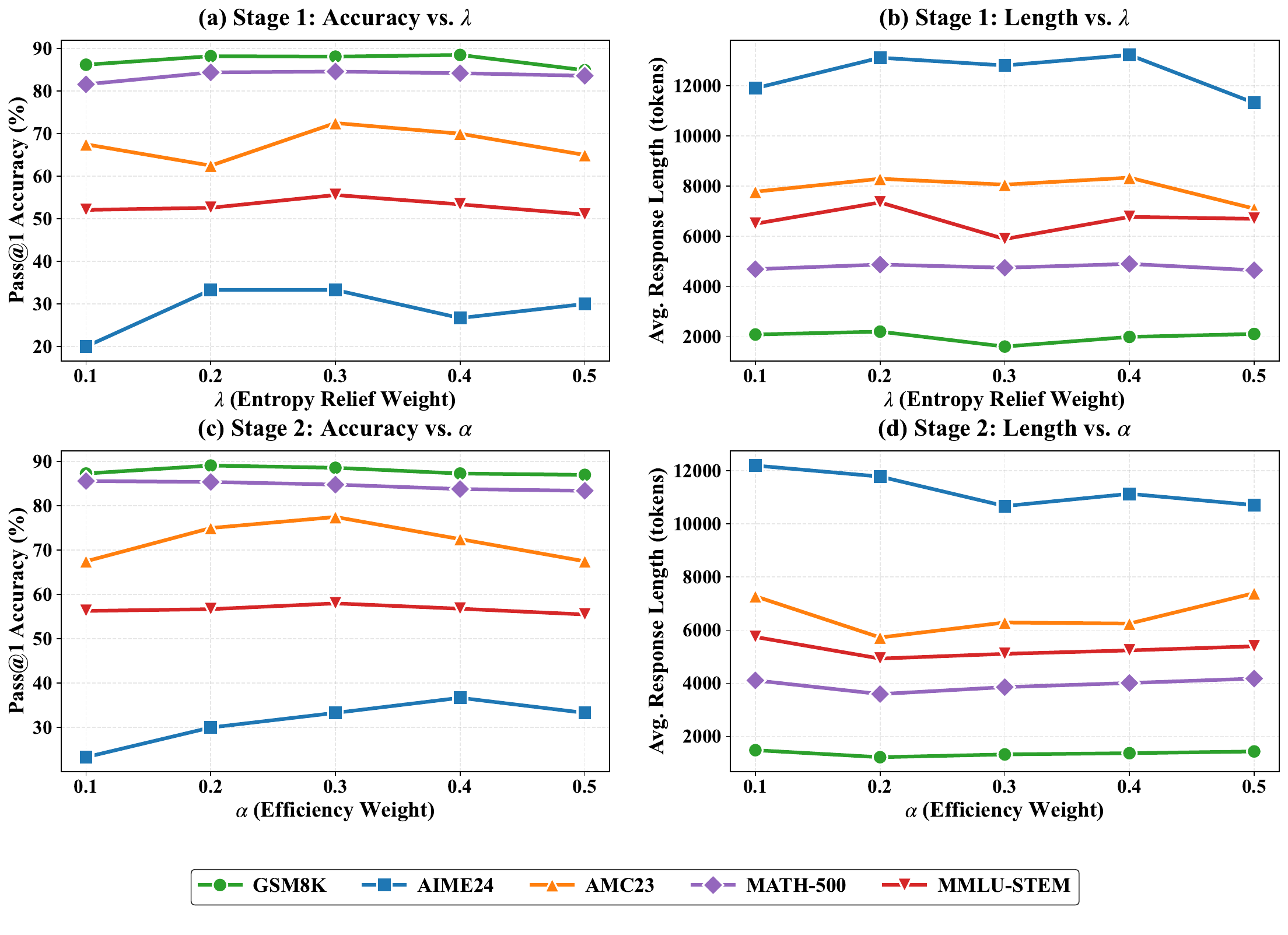}
  \caption{\textbf{Hyperparameter sensitivity analysis across five
  benchmarks.} We vary the entropy relief weight $\lambda$ (Phase 1,
  rows) and efficiency weight $\alpha$ (Phase 2, columns) on
  \texttt{DeepSeek-R1-Distill-Qwen-1.5B}. Each heatmap shows accuracy
  (Acc) and token count (Tok). The default values $\lambda=0.3$ and
  $\alpha=0.3$ consistently yield the best accuracy--length trade-off
  across all five datasets, with marginal degradation toward the
  extremes.}
  \label{fig:hyperparam}
\end{figure*}

We perform a comprehensive sensitivity analysis for the two main
hyperparameters controlling the magnitude of reward bonuses: $\lambda$
(entropy relief weight in phase-1) and $\alpha$ (efficiency weight in
phase-2). Figure~\ref{fig:hyperparam} visualizes the accuracy--length
trade-off across all five benchmarks as we vary $\lambda \in
\{0.1, 0.2, 0.3, 0.4, 0.5\}$ and $\alpha \in \{0.1, 0.2, 0.3, 0.4,
0.5\}$.

\paragraph{Phase-1 sensitivity ($\lambda$).}
Lower values ($\lambda=0.1$) weaken the entropy-drop reward signal,
resulting in reduced accuracy. As $\lambda$ increases from 0.1 to 0.3,
accuracy improves progressively. Beyond $\lambda=0.3$, accuracy plateaus
or degrades slightly, suggesting over-emphasis on entropy drops biases
the model toward verbose justifications. The log-normalization in
Eq.~\eqref{eq:err_score} provides natural regularization, but excessive
weight ($\lambda \geq 0.4$) can overwhelm it.

\paragraph{Phase-2 sensitivity ($\alpha$).}
Lower values ($\alpha=0.1$) apply gentle length pressure but without
meaningful compression. Increasing $\alpha$ to 0.2 and 0.3 progressively
improves the accuracy--length trade-off. Beyond $\alpha=0.3$, compression
saturates and accuracy begins to degrade. The $\tanh$ saturation
in Eq.~\eqref{eq:lambda} prevents runaway penalization at extreme
$z$-scores, but excessive $\alpha$ shifts the gradient balance too heavily
toward length reduction, sacrificing reasoning quality.

\paragraph{Cross-benchmark consistency.}
Across all five benchmarks (visualized in Figure~\ref{fig:hyperparam}),
the default values $\lambda=0.3$ and $\alpha=0.3$ consistently yield the
best accuracy--length trade-off. GSM8K (easier, grade-school math) shows
less sensitivity to these parameters, while AIME24 and AMC23 (harder,
competition math) benefit more from careful tuning. This suggests that
on harder reasoning tasks, the distinction between encouraging entropy
drops and penalizing length becomes more critical. The log-normalization
and $\tanh$ saturation provide robustness across problem difficulty levels.

\paragraph{Robustness region.}
The region $\lambda \in [0.2, 0.4]$ and $\alpha \in [0.2, 0.4]$ represents
a ``sweet spot'' where \method{} maintains $>87\%$ average accuracy while
achieving $>18\%$ length compression. Values outside this region show
either accuracy degradation (too low: insufficient signal) or inefficient
length compression (too high: conflicting objectives). Our default
$\lambda=0.3, \alpha=0.3$ sits at the center of this region, providing
balanced improvement.

\section{Prompt Format}
\label{app:prompt}

In our experiments, when training models on the GSM8K dataset~\citep{cobbe2021gsm8k}, we uniformly apply the following prompt to elicit step-by-step reasoning and format the final output:

\begin{tcolorbox}[colback=white, colframe=black, boxrule=0.5pt,
  arc=0pt, left=4pt, right=4pt, top=4pt, bottom=4pt]
\small
\begin{verbatim}
Let's think step by step and output the final
answer after "####".
\end{verbatim}
\end{tcolorbox}

During the evaluation phase, we follow the official protocol of 
Qwen2.5-Math (\url{https://github.com/QwenLM/Qwen2.5-Math}). We 
employ their specific chat template and system instruction to guide 
the model's step-by-step reasoning and standardize the final answer 
delimiter. The exact prompt template is:

\begin{tcolorbox}[colback=white, colframe=black, boxrule=0.5pt,
  arc=0pt, left=4pt, right=4pt, top=4pt, bottom=4pt]
\small
\begin{verbatim}
"qwen25-math-cot": (
    "<|im_start|>system\n
    Please reason step by step, and put your final
    answer within \\boxed{{}}.<|im_end|>\n"
    "<|im_start|>user\n{input}<|im_end|>\n"
    "<|im_start|>assistant\n",
    "{output}",
    "\n\n",
)
\end{verbatim}
\end{tcolorbox}

\begin{figure*}[t]
  \centering
  \includegraphics[width=0.94\textwidth]{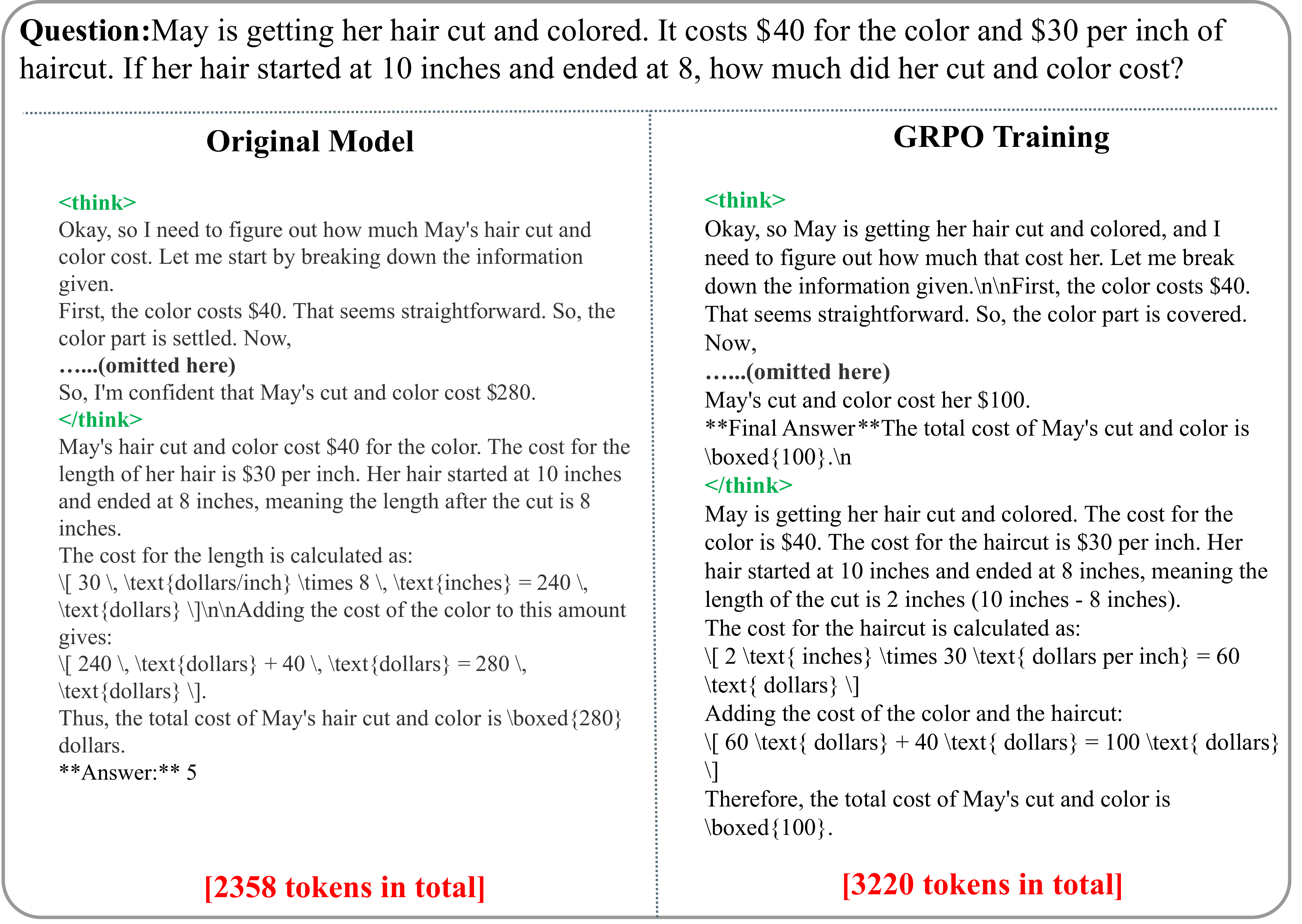}
  \caption{%
    \textbf{Original model vs.\ GRPO.} The original model (left,
    2358 tokens) produces the wrong answer; GRPO (right, 3220 tokens)
    corrects it but at the cost of increased response length.
  }
  \label{fig:case_phase1}
\end{figure*}

\begin{figure*}[t]
  \centering
  \includegraphics[width=0.94\textwidth]{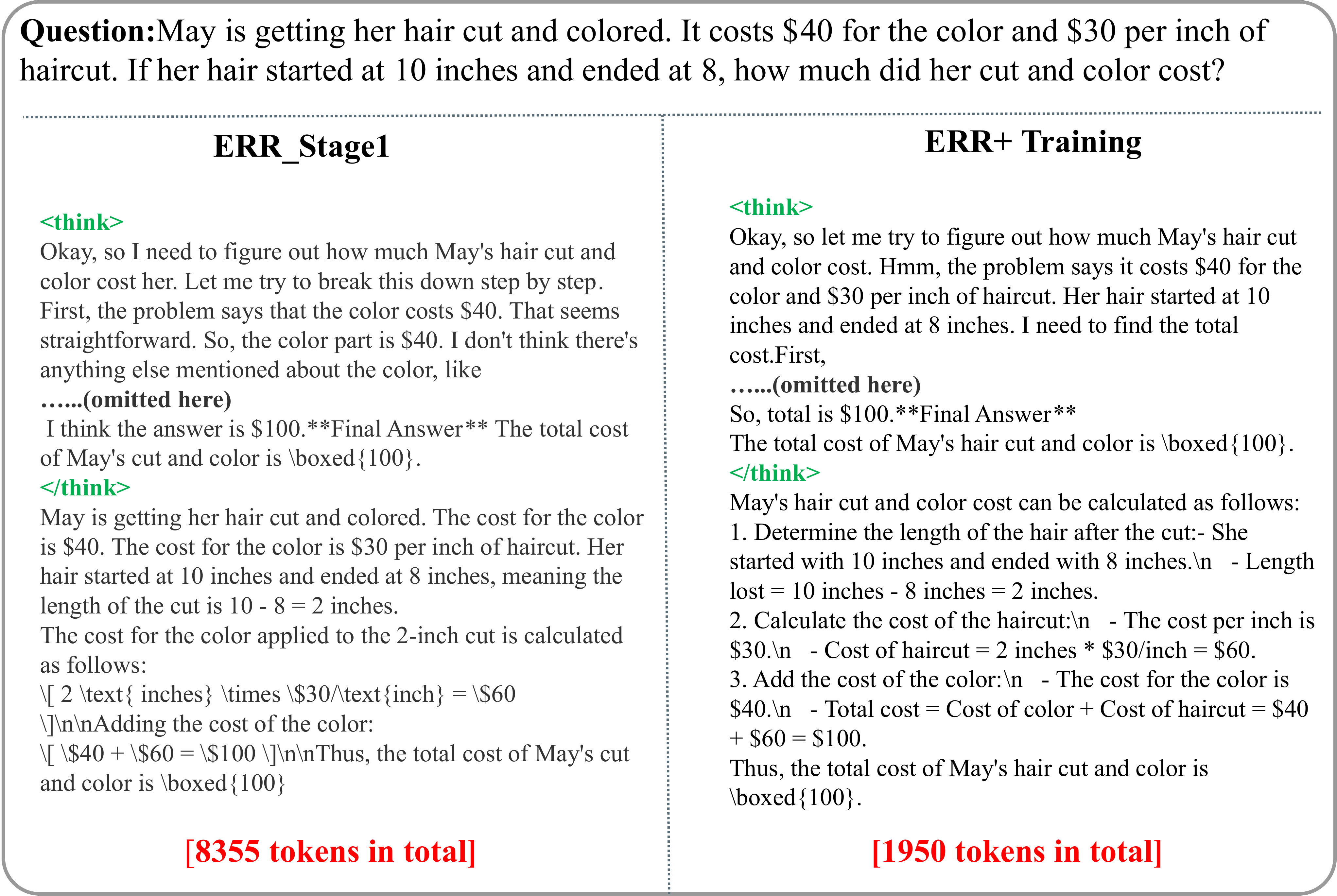}
  \caption{%
    \textbf{\methodbase{} vs.\ \method{}.} Both reach the correct
    answer; \method{} (right, 1950 tokens) reduces response length
    by 77\% compared to \methodbase{} (left, 8355 tokens).
  }
  \label{fig:case_length}
\end{figure*}

\section{Case Studies}
\label{app:cases}

We present a representative GSM8K problem traced across four
checkpoints to illustrate the progressive effect of each training
phase. Figure~\ref{fig:case_phase1} compares the original model and
GRPO training. The original model produces 2358 tokens but arrives
at the wrong answer (\$280), misinterpreting the haircut length.
GRPO corrects this but increases token usage to 3220 tokens,
confirming that binary correctness rewards provide no conciseness
signal. Figure~\ref{fig:case_length} shows that \methodbase{}
further corrects reasoning structure while \method{} additionally
compresses the response from 8355 to 1950 tokens (77\% reduction),
committing directly to each arithmetic step without redundant
elaboration. Taken together, the two figures confirm that Phase~1
establishes correct reasoning structure and Phase~2 safely compresses
length without disrupting it.

\section{Statistical Performance Analysis of Different Methods}

\begin{figure*}[t]
  \centering
  \includegraphics[width=0.94\textwidth]{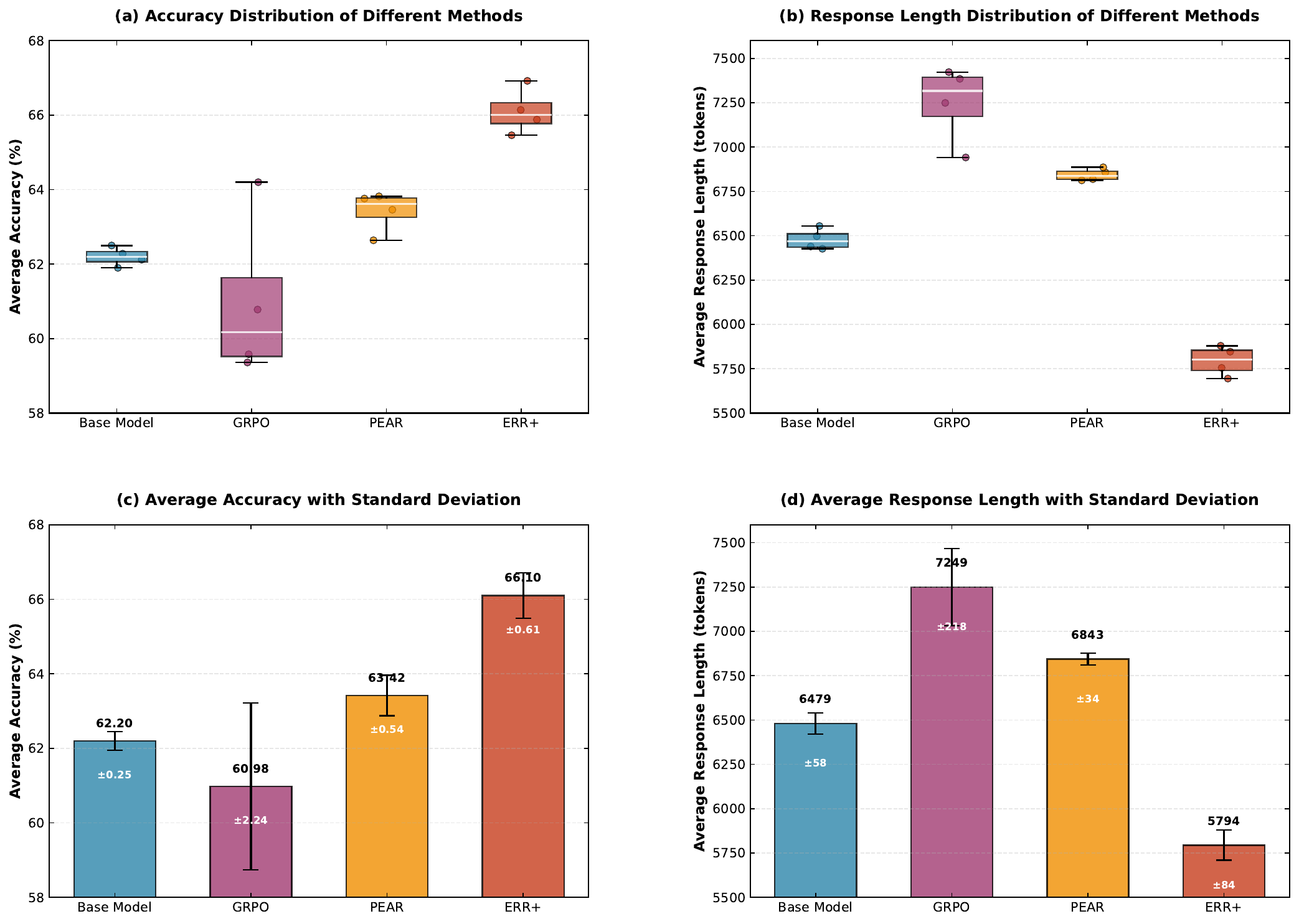}
  \caption{Performance and Stability Analysis of Different Methods for DeepSeek-R1-Distill-Qwen-1.5B}
  \label{fig:stability_analysis}
\end{figure*}

As Figure~\ref{fig:stability_analysis} shows, to measure the stability of different methods, we computed the average accuracy and response length of the DeepSeek-R1-Distill-Qwen-1.5B (DS-R1-1.5B) model across five distinct datasets, based on four independent runs for each approach. From a statistical perspective, the ERR+ method demonstrates remarkable comprehensive advantages on the DS-R1-1.5B model. Its accuracy superiority has been verified by rigorous t-tests with $p$-values all below 0.05 in comparisons with other methods, confirming clear statistical significance and effectively ruling out the impact of random errors. Meanwhile, the method exhibits excellent stability and reliability: an accuracy standard deviation of 0.61\% and variance of 0.3773 reflect high consistency across multiple independent evaluations, ensuring its stability fully meets practical application requirements. Furthermore, it delivers notable efficiency superiority: while maintaining the highest accuracy among all methods, it achieves the shortest response length of 5794 tokens, with a response length standard deviation of only 84 tokens that guarantees controllable fluctuations, thus perfectly balancing inference efficiency and result stability. In comparison with traditional RL methods such as GRPO and PEAR, the ERR+ method also presents distinct advantages---it not only mitigates the poor stability issue of GRPO and overcomes the insufficient accuracy limitation of PEAR, but also achieves a breakthrough improvement in output efficiency.

\section{Trace Structure Analysis: Why Phase~1 Enables Phase~2}
\label{app:structure}

Section~\ref{sec:theory} provides a formal gradient-conflict argument for the
sequential design. Here we offer complementary empirical evidence at the trace
level, explaining the mechanistic link between ERR and length efficiency.

\paragraph{Definitions.}
We call a position a \emph{large drop} when the entropy relief
$r_t > \tau$ (threshold $\tau = 0.5$ nats).
A \emph{low-relief span} is a contiguous segment of thinking tokens
containing no large drop.
Large drops are \emph{commitment points} where the model resolves a
sub-problem; low-relief spans are the explanatory tokens between commitments.

\paragraph{Mechanism diagnostics.}
To further validate the claims that (i)~phase~1 promotes
uncertainty resolution without collapsing exploration and
(ii)~gradient conflict decreases after phase~1 convergence,
we report four additional statistics at three checkpoints in
Table~\ref{tab:mechanism}: average thinking-phase entropy
($\bar{\entropy}$), the mid-exploration fraction $\rho$
(high-entropy tokens not yet followed by a committed drop,
as defined in Section~\ref{sec:theory}), and the cosine
similarity between $g_1$ and $g_2$.

\begin{table}[h]
\centering
\small
\caption{Mechanism diagnostics on GSM8K
  (\texttt{DeepSeek-R1-Distill-Qwen-1.5B}).
  $\bar{\mathcal{H}}$: mean thinking-phase entropy.
  $\rho$: mid-exploration fraction.
  Grad cos: cosine between $g_1$ and $g_2$.}
\label{tab:mechanism}
\setlength{\tabcolsep}{0pt}
\begin{tabular*}{\columnwidth}{@{\extracolsep{\fill}}lccccc@{}}
\toprule
\textbf{Stage} & \textbf{Acc} & \textbf{Tok}
  & $\bar{\mathcal{H}}$ & $\rho$ & \textbf{Grad cos} \\
\midrule
Before ERR      & 84.6 & 2076 & 0.58 & 0.28 & $-$0.45 \\
After ERR (Ph.1)& 88.1 & 1612 & 0.49 & 0.08 & $-$0.05 \\
After \method{} & 88.6 & 1324 & 0.52 & 0.06 & $+$0.02 \\
\bottomrule
\end{tabular*}
\end{table}

Phase~1 doubles the large-drop frequency and reduces $\rho$
from 0.28 to 0.08, confirming that high-entropy positions are
increasingly resolved into committed drops rather than
remaining as unresolved exploration.
Average entropy does not collapse ($0.58\to0.49$), consistent
with the claim that \methodbase{} preserves exploratory states.
The gradient cosine rises from $-0.45$ to near zero after
phase~1 and turns slightly positive after phase~2,
empirically confirming Theorem~\ref{thm:conflict}'s prediction
that the conflict dissolves at phase~1 convergence.

\paragraph{Safe-prune accuracy: formal definition and protocol.}
Safe-prune accuracy is the percentage of \emph{originally correct}
samples whose extracted final answers remain correct when the model is
forced to continue from a truncated trajectory with low-relief spans
removed. It evaluates token-level continuation from the pruned state,
not a restart of reasoning from scratch. Concretely: we filter the
evaluation logs (\texttt{.jsonl}) for traces with
\texttt{score[0]==True}; for each training stage we sample 200 correct
GSM8K instances as the denominator. For each instance we remove
low-relief spans (flat-entropy regions $>20$ tokens between adjacent
large-drop commitment points) and replace them with ``\texttt{...}''.
The model is fed this truncated prefix in the original dialogue
template and prompted for continuation (greedy decoding up to 128 tokens
until \texttt{Final Answer:}). If the regenerated answer satisfies
\texttt{math\_equal(pred, gt)} it counts toward the numerator. Grad cos
is the cosine similarity between accumulated gradients $g_{\mathrm{ERR}}$
and $g_{\mathrm{RRER}}$ over sampled trajectories, estimated over the
full 1319-sample GSM8K test set.

\paragraph{Stability on small benchmarks.}
We further verify these metrics on AIME24 and AMC23 with
\texttt{DeepSeek-R1-Distill-Qwen-1.5B} over four independent runs
($T{=}0.6$, top-$p{=}0.95$). Safe-prune is computed on originally
correct logs (8 instances for AIME24, 28 for AMC23); Grad cos is
estimated over the full benchmarks. Error bars are the unbiased sample
standard deviation (Bessel's correction, $N-1$). The cross-checkpoint
trends generalize beyond GSM8K: safe-prune accuracy consistently rises
after ERR, and Grad cos shifts from large negative conflict toward
near-zero alignment after Phase~1 and \method{}.

\begin{table}[t]
\centering
\small
\setlength{\tabcolsep}{4pt}
\caption{Safe-prune accuracy and gradient cosine on small benchmarks
  over four runs (\texttt{DeepSeek-R1-Distill-Qwen-1.5B}).}
\label{tab:small_stability}
\resizebox{\columnwidth}{!}{%
\begin{tabular}{llcc}
\toprule
\textbf{Bench.} & \textbf{Stage} & \textbf{Safe-prune (\%)} & \textbf{Grad cos} \\
\midrule
\multirow{3}{*}{AIME24}
  & Before ERR       & $65.63 \pm 6.25$  & $-0.78 \pm 0.04$ \\
  & After ERR (Ph.1) & $75.00 \pm 10.21$ & $-0.26 \pm 0.02$ \\
  & After \method{}  & $78.13 \pm 6.25$  & $-0.02 \pm 0.02$ \\
\midrule
\multirow{3}{*}{AMC23}
  & Before ERR       & $75.90 \pm 3.45$  & $-0.56 \pm 0.03$ \\
  & After ERR (Ph.1) & $82.13 \pm 2.90$  & $-0.12 \pm 0.02$ \\
  & After \method{}  & $83.90 \pm 2.08$  & $+0.01 \pm 0.02$ \\
\bottomrule
\end{tabular}%
}
\end{table}

\paragraph{Structure-then-compress mechanism.}
Table~\ref{tab:structure} measures four statistics at three checkpoints
on GSM8K (\texttt{DeepSeek-R1-Distill-Qwen-1.5B}):
(i)~\emph{large-drop frequency} (large drops per 100 thinking tokens),
(ii)~\emph{gap} (average tokens between adjacent large drops),
(iii)~\emph{low-relief span \%} (thinking tokens inside no-large-drop segments),
and (iv)~\emph{safe-prune accuracy} (Pass@1 after removing all low-relief spans).

\begin{table}[t]
\centering
\small
\caption{Trace structure statistics at three checkpoints on GSM8K.
  \emph{Safe-prune acc.}\ measures Pass@1 after removing all low-relief spans.}
\label{tab:structure}
\setlength{\tabcolsep}{0pt}  % 关键：清零默认列间距
\begin{tabular*}{\columnwidth}{@{\extracolsep{\fill}}lcccc@{}}
\toprule
\textbf{Stage} & \textbf{Drop freq.} & \textbf{Gap} & \textbf{Low-relief \%} & \textbf{Safe-prune acc.} \\
\midrule
Before ERR       & 6.5  & 15.4 & 32.5 & 72.4 \\
After ERR        & 12.8 & 7.8  & 35.8 & 84.3 \\
After \method{}  & 14.2 & 7.0  & 24.5 & 84.8 \\
\bottomrule
\end{tabular*}
\end{table}

Phase~1 doubles the large-drop frequency (6.5$\to$12.8) and raises
safe-prune accuracy by 12 points (72.4$\to$84.3\%), confirming that
traces are now organized around clearly identifiable commitment points.
Phase~2 then reduces the low-relief span percentage substantially
(35.8$\to$24.5\%) without hurting safe-prune accuracy (84.3$\to$84.8\%),
confirming that it compresses exactly the redundant material that Phase~1
has rendered safe to remove.
This \emph{structure-then-compress} mechanism directly explains why ERR
and \methodlen{} form a cohesive two-phase design: Phase~1 must first
consolidate reasoning into discrete commitments before Phase~2 can safely
eliminate the low-value spans between them.

\begin{table}[h]
\centering
\footnotesize
\caption{Results of DAPO and DAPO+\method{} on \texttt{DeepSeek-R1-Distill-Qwen-1.5B}
across four benchmarks, reporting Pass@1 accuracy (\%) and average response length
(tokens). \method{} reward signals consistently improve both metrics over the DAPO
baseline, demonstrating generalization beyond GRPO.}
\label{tab:dapo}
\resizebox{\columnwidth}{!}{%
\begin{tabular}{l cc cc cc cc}
\toprule
\multirow{2}{*}{\textbf{Method}}
  & \multicolumn{2}{c}{\textbf{GSM8K}}
  & \multicolumn{2}{c}{\textbf{AIME24}}
  & \multicolumn{2}{c}{\textbf{AMC23}}
  & \multicolumn{2}{c}{\textbf{MATH-500}} \\
\cmidrule(lr){2-3}\cmidrule(lr){4-5}\cmidrule(lr){6-7}\cmidrule(lr){8-9}
  & Acc & Tok & Acc & Tok & Acc & Tok & Acc & Tok \\
\midrule
DAPO
  & 85.4 & 2123 & 26.7 & 12956 & 70.0 & 8065 & 83.4 & 4432 \\
DAPO+\method{}
  & \textbf{87.1} & \textbf{1465}
  & \textbf{30.0} & \textbf{10542}
  & \textbf{75.0} & \textbf{5603}
  & \textbf{85.2} & \textbf{3296} \\
\bottomrule
\end{tabular}%
}
\end{table}

\section{Generalization to DAPO}
\label{app:dapo}

To verify that \method{} is not specific to GRPO, we integrate the two-phase reward
design with DAPO~\citep{yu2025dapo} as the base optimizer, training on
\texttt{DeepSeek-R1-Distill-Qwen-1.5B} under identical data and evaluation settings.
Table~\ref{tab:dapo} shows that DAPO+\method{} consistently improves over
DAPO alone in both accuracy and response length across four benchmarks,
demonstrating that the ERR and \methodlen{} reward signals transfer across
different policy optimization backbones.

\section{Generalization to Non-Math Reasoning}
\label{app:gpqa}
To test whether the entropy-drop pattern extends beyond
arithmetic-style tasks, Table~\ref{tab:scope_generality}
reports results on GPQA Diamond~\citep{Rein2023GPQAAG}, a
graduate-level benchmark spanning biology, physics, and
chemistry, alongside MMLU-STEM for reference.
Crucially, GPQA Diamond and MMLU-STEM are evaluated \emph{directly after
GSM8K-only training, without any cross-domain fine-tuning}, making this
a genuine cross-domain stress test.
\method{} improves both accuracy and response length over
GRPO on both benchmarks, indicating that the model acquires a general
reasoning-efficiency behavior rather than overfitting to GSM8K-specific
format or arithmetic patterns, and that entropy relief
is not specific to step-by-step arithmetic reasoning.

\end{document}